\pdfoutput=1

\PassOptionsToPackage{dvipsnames}{xcolor}

\documentclass[11pt]{article}

\usepackage{EMNLP2023}

\usepackage{times}
\usepackage{latexsym}

\usepackage[T1]{fontenc}

\usepackage[utf8]{inputenc}

\usepackage{microtype}

\usepackage{inconsolata}

\usepackage{amsmath}
\usepackage{graphicx}
\usepackage{color}
\usepackage{xcolor}
\usepackage{amsfonts}
\usepackage{latexsym}
\usepackage{url}
\usepackage{wrapfig}
\usepackage{comment}
\usepackage{booktabs}
\usepackage{microtype}
\usepackage{xspace}
\usepackage{booktabs}
\usepackage{multirow}
\usepackage{dcolumn}
\usepackage{hhline}
\usepackage{mdframed}
\usepackage{enumitem}
\usepackage{xcolor}
\usepackage{csquotes}
\newcolumntype{d}[1]{D{.}{.}{#1}}
\usepackage{subcaption}
\usepackage{makecell}
\usepackage{color,soul}
\usepackage{algorithm}
\usepackage[noend]{algpseudocode}
\usepackage{xcolor,colortbl}
\usepackage{xspace}
\usepackage{mdframed}
\usepackage[outline]{contour}
\usepackage{hyperref}

\newmdtheoremenv{mydef}{Factual Precision}

\makeatletter
\newcommand\footnoteref[1]{\protected@xdef\@thefnmark{\ref{#1}}\@footnotemark}
\makeatother

\definecolor{green}{rgb}{0.1,0.1,0.1}
\definecolor{chocolate}{HTML}{D2691E}
\definecolor{maroon}{HTML}{A00000}
\definecolor{indigo}{HTML}{4B0082}
\definecolor{green}{HTML}{008000}
\definecolor{red}{HTML}{a91e1e}
\definecolor{cadmiumgreen}{rgb}{0.0, 0.42, 0.24}

\newcommand{\best}[1]{\textcolor{red}{\textbf{#1}}}

\newcolumntype{L}[1]{>{\PreserveBackslash\raggedright}p{#1}}
\newcolumntype{R}[1]{>{\raggedleft\let\newline\\\arraybackslash\hspace{0pt}}m{#1}}

\usepackage{amssymb}
\usepackage{pifont}
\newcommand{\cmark}{{\protect\color{maroon} \ding{51}}}
\newcommand{\xmark}{\ding{55}}
\newcommand{\gmark}{{\protect\color{ForestGreen} \ding{51}}}
\makeatletter
\newcommand*\myfontsize{%
  \@setfontsize\myfontsize{8}{9}%
}
\makeatother

\makeatletter
\newcommand*\mysmallfontsize{%
  \@setfontsize\mysmallfontsize{7.4}{8.4}%
}
\makeatother

\newcommand{\ours}{\textsc{FActScore}}

\newcommand{\ourslong}{\textbf{F}actual precision in \textbf{A}tomi\textbf{c}i\textbf{t}y \textbf{Score}}
\newcommand{\ourScore}{\textsc{FActScore}}
\newcommand{\ourScores}{\textsc{FActScore}s}

\newcommand{\subjectLM}{LM$_\textsc{subj}$}
\newcommand{\evalLM}{LM$_\textsc{eval}$}

\newcommand{\sLabel}{\texttt{Supported}}
\newcommand{\nsLabel}{\texttt{Not-supported}}
\newcommand{\irLabel}{\texttt{Irrelevant}}

\newcommand{\nsLabelShort}{\texttt{NS}}

\newcommand{\rtg}{Retrieve$\rightarrow$LM}
\newcommand{\rtgShort}{Retrv$\rightarrow$LM}

\newcommand{\Fone}{F1$_\textsc{micro}$}

\usepackage{adjustbox}
\newcommand{\marktext}[2]{\adjustbox{bgcolor=#1}{\strut #2}}
\newcommand{\myblue}[1]{\marktext{blue!30}{\texttt{#1}}}
\newcommand{\mypink}[1]{\marktext{pink!70}{\texttt{#1}}}
\newcommand{\mypurple}[1]{\marktext{purple!30}{\texttt{#1}}}

\newcommand{\myskip}[1]{}

\newcommand{\alwaysS}{ 30.8 & 37.1 & 45.0 }
\newcommand{\alwaysNS}{ 35.7 & 29.1 & 15.5 }
\newcommand{\alwaysRandom}{ 50.5 & 50.2 & 43.2 }

\newcommand{\noContextQA}{ 56.5 & 48.8 & 32.5 }
\newcommand{\noContextFV}{ 57.3 & 55.3 & 41.7 }
\newcommand{\selfcheckQA}{ 65.3 & 63.2 & - }
\newcommand{\selfcheckFV}{ 68.0 & 61.9 & - }

\newcommand{\pw}[1]{
}

\title{
    \ours:
    Fine-grained Atomic Evaluation of \\ Factual Precision in Long Form Text Generation

}

\newcommand{\affilsup}[1]{\rlap{\textsuperscript{\normalfont#1}}}

\author{Sewon Min\affilsup{$\dagger$1} \quad
    Kalpesh Krishna\affilsup{$\dagger$2} \quad
    Xinxi Lyu\affilsup{1} \quad
    \textbf{Mike Lewis}\affilsup{4} \quad 
    \textbf{Wen-tau Yih}\affilsup{4} \\
    \textbf{Pang Wei Koh}\affilsup{1} \quad
    \textbf{Mohit Iyyer}\affilsup{2} \quad
    \textbf{Luke Zettlemoyer}\affilsup{1,4} \quad
    \textbf{Hannaneh Hajishirzi}\affilsup{1,3} \\
    $^1$University of Washington \quad
    $^2$University of Massachusetts Amherst \\
    $^3$Allen Institute for AI \quad
    $^4$Meta AI \\
    \texttt{\{sewon,alrope,pangwei,lsz,hannaneh\}@cs.washington.edu} \\
    \texttt{\{kalpesh,miyyer\}@cs.umass.edu} \quad
    \texttt{\{mikelewis,scottyih\}@meta.com}
}

\begin{document}
\maketitle

\def\thefootnote{$\dagger$}\footnotetext{Core contributors.}\def\thefootnote{\arabic{footnote}}

\begin{abstract}
    Evaluating the factuality of long-form text generated by large language models (LMs) is non-trivial because (1) generations often contain a mixture of supported and unsupported pieces of information, making binary judgments of quality inadequate, and (2) human evaluation is time-consuming and costly.
    In this paper, we introduce \textbf{\ours}, 
    a new evaluation 
    that breaks a generation into a series of atomic facts and computes the percentage of atomic facts supported by a reliable knowledge source.
    We conduct an extensive human evaluation to obtain \ourScores\ of people biographies generated by several state-of-the-art commercial LMs---InstructGPT, ChatGPT, and the retrieval-augmented PerplexityAI---and report new analysis demonstrating the need for such a fine-grained score (e.g., ChatGPT only achieves 58\%).
    Since human evaluation is costly, 
    we also introduce an automated model that estimates \ours\ using retrieval and a strong language model, with less than a $2\%$ error rate. 
    Finally, we use this automated metric to evaluate 6,500 generations from a new set of 13 recent LMs that would have cost \$26K if evaluated by humans, with various findings: GPT-4 and ChatGPT are more factual than public models, and Vicuna and Alpaca are some of the best public models. \ours\ is available for public use via \texttt{pip install factscore}.\footnote{Source code and guidelines are available at \url{https://github.com/shmsw25/FActScore}.}
\end{abstract}

\section{Introduction}\label{sec:intro}Long-form text generated by large language models (LMs) has widely been used~\citep{gpt3,ouyang2022training};
nonetheless, evaluating their {\em factual precision}---whether each piece of information conveyed in a generation is factually accurate---remains challenging for two reasons.
First, a generation consists of a large number of pieces of information that are a mixture of true or false,\footnote{
Even a single sentence consists of multiple pieces of information (e.g., 4.4 per sentence in ChatGPT, 40\% of which are a mixture of supported and unsupported information).} making a binary judgment inadequate~\citep{pagnoni-etal-2021-understanding}.
Second, validating every piece of information is time-consuming and costly.

\begin{figure}[t]
\centering \footnotesize
\resizebox{\columnwidth}{!}{
    \includegraphics[trim={8.2cm 8cm 14cm 1cm},clip]{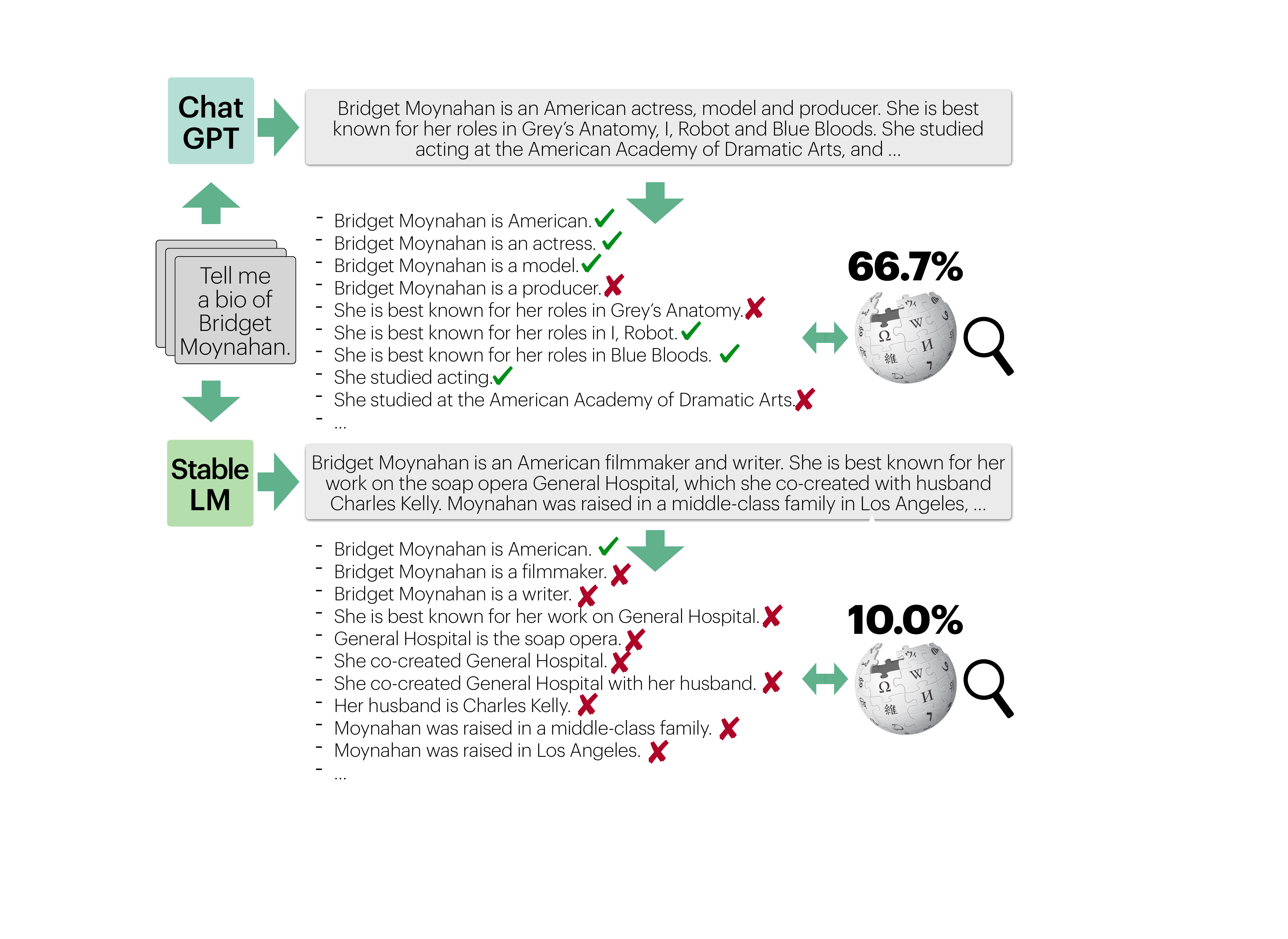}%
}\vspace{-.3em}
\caption{
    An overview of \ours, a fraction of \emph{atomic facts} (pieces of information) supported by a given knowledge source.
    %
    \ours\ allows a more fine-grained evaluation of factual precision, e.g., in the figure, the top model gets a score of 66.7\% and the bottom model gets 10.0\%, whereas prior work would assign 0.0 to both.
    \ours\ can either be based on human evaluation, or be automated, which allows evaluation of a large set of LMs with no human efforts.
}\label{fig:teaser}
\end{figure}

In this paper, we introduce \textbf{\ours} (\ourslong), a new evaluation of an LM that represents {\em the percentage of atomic facts (pieces of information) supported by a given knowledge source}.
%
Computing \ours\ involves
(1) breaking a generation into a series of atomic facts---short statements that each contain one piece of information~\citep{nenkova2004evaluating,shapira-etal-2019-crowdsourcing,zhang-bansal-2021-finding,liu2022revisiting}, and (2) assigning a binary label 
to each atomic fact, allowing a fine-grained evaluation of factual precision.
%
We evaluate \ours\ on the task of generating people biographies because
generations
consist of verifiable statements rather than debatable or subjective ones,
and the scope is broad
(i.e., covering diverse nationalities, professions, and levels of rarity). 

%
We perform extensive human annotations to obtain \ourScores\ of three state-of-the-art, commercially available LMs: InstructGPT~\citep{ouyang2022training}, ChatGPT~\citep{chatgpt}, and search-augmented PerplexityAI.\footnote{\label{fn:ppl}\href{https://www.perplexity.ai/}{\texttt{perplexity.ai}}}
Our results indicate that commercially available LMs are riddled with errors, having \ourScores\ of 42\%, 58\% and 71\%, respectively.
Their \ourScores\ significantly drop as the rarity of the entities increases, e.g., $80\%\rightarrow16\%$ for ChatGPT.

Since human evaluation is costly, we next introduce an automatic evaluation of \ours\ through a model that estimates a \ourScore\ for a given LM.
Our estimator decomposes generations into atomic facts and validates each based on a given knowledge source, leveraging retrieval from the given knowledge source and strong language models.
Our estimator closely approximates \ours\ with an error rate of $<2\%$
%
and can be applied to a range of {\em new} LMs at scale with no human effort.
Our case study evaluates 6,500 generations from 13 LMs that could have cost \$26K, 
with various findings: 
GPT-4~\citep{OpenAI2023GPT4TR} and ChatGPT are far less factual than humans 
but are much better than public models,
and there is a large variance between public models, with 
Vicuna~\citep{vicuna2023} and Alpaca~\citep{alpaca} being some of the best.


In summary, our contributions are as follows.
\vspace{-.3em}
\begin{enumerate}[leftmargin=15pt]\itemsep -.2em
    \item We introduce \ours, a new evaluation of factual precision of LMs by breaking their generations into atomic facts and validating each against a given knowledge source. Human evaluation reveals that the state-of-the-art LMs with and without search have low \ourScores. 
    \item We introduce a model that approximates \ours\ with an error rate of $<2\%$, allowing evaluation of a large set of new LMs without manual human efforts.
    \item We open-sourced \ours\ and the annotated data for public use, available via \texttt{pip install factscore}.
    We suggest future work to extend \ours\ for a broader set of generations (e.g., open-ended generation) and to further improve the estimator.
\end{enumerate}
\pw{Overall, I think the introduction could frame this paper more clearly in the context of prior work: what is the key idea and how is it different from prior work? And what are the results/analysis/insights that this idea enables?}
\section{Related Work}\label{sec:related}\paragraph{Factual precision in text generation.}

Factual precision in text generation 
has been an active area of research in NLP. 
Most prior work studies factual precision of models supervised for a specific problem such as dialogue~\citep{shuster2021retrieval}, or focuses on question answering with short answers~\citep{kadavath2022language,kandpal2022large,mallen2022not,nori2023capabilities}.

More recent work has studied factual precision of text generation beyond short answers. 
\citet{lee2022factuality} evaluates the factual precision with proxy metrics, e.g., whether named entities in a generation appear in an article of the topic.
A series of concurrent work verifies the precision of the citations (attributions) provided by the model~\citep{gao2022attributed,liu2023evaluating,yue2023automatic,gao2023enabling}.
A concurrent work by \citet{manakul2023selfcheckgpt} automates the identification of factual errors in LM generations without using any knowledge source; we use their method as a baseline estimator in Section~\ref{sec:method}.
In contrast, our work (1) considers much longer text generation\footnote{Consisting of 110--151 words (Table~\ref{tab:longform-data-statistics}), in contrast to 18--29 in \citet{gao2022attributed} and 65 in \citet{liu2023evaluating}.} from a variety of state-of-the-art LMs with and without search, (2) provides their fine-grained evaluation both by human experts and through an automated evaluator that closely approaches humans, and (3) applies it to a large set of LMs at scale.

\myskip{
More recent work has studied factual precision in the LM-generated text, as summarized in Table~\ref{tab:longform-related}. 
They include concurrent work from \citet{manakul2023selfcheckgpt} which proposes to validate factuality of generation without a nonparametric component, which we compare against in Section~\ref{subsec:models-validation}, and \citet{liu2023evaluating}, which evaluates veracity of citations provided by commercial generative search engines.
Our work is the first that (1) evaluates \textbf{\em long}-form generation from a variety of state-of-the-art base models with and without search, (2) provides fine-grained labels in precision (three-level labels that consider two different causes of precision errors, atomic-level instead of sentence-level; detailed in Section~\ref{sec:benchmark}), and (3) benchmarks a range of verification models, highlighting the importance of each component.
}

\vspace{-.3em}
\paragraph{Fact Verification.} Our work is closely related to prior work on fact verification~\citep{thorne-etal-2018-fever,wadden-etal-2020-fact} where claim sentences are automatically checked against a large knowledge source like Wikipedia or scientific literature.
Most literature assumes a single, atomic claim, sometimes modeled with surrounding context~\citep{nakov2018overview, mihaylova2019semeval, shaar-etal-2022-role}.
There also has been work that verifies a longer sentence or text through decomposition to atomic facts~\citep{fan2020generating,wright-etal-2022-generating,chen2022generating,kamoi2023wice}
from which we take inspiration.
The primary difference between fact verification literature and our work is that we focus on long-form {\em model-generated} text rather than sentence-level human-written claims.

\vspace{-.3em}
\paragraph{Model-based Evaluation.}
Prior work has used learned models to define automated evaluation scores~\citep{zhang2019bertscore,liu2023gpteval}.
This includes model-based evaluation in summarization that considers the consistency between a summary and a source document using QA or NLI~\citep{kryscinski2019evaluating,wang2020asking,fabbri2021qafacteval,deutsch2021towards,laban2022summac}.
We take inspiration from this work, and evaluate factual precision of LM generations by considering whether pieces of information are supported by a large text corpus.

\section{\ours: Evaluating Factual Precision of Long-form Text Generation}\label{sec:benchmark}

We introduce \ours, a new evaluation of an LM that considers the factual precision of atomic facts generated by the LM.
We perform human evaluations to calculate \ourScores\ of the state-of-the-art LMs (Section~\ref{subsec:data}) and discuss results (Section~\ref{subsec:annotation-results}).
\ours\ allows rigorous and fine-grained evaluation of factual precision, but is time-consuming and costly, motivating automatic evaluation in Section~\ref{sec:method}.

\subsection{Definition}
\label{subsec:data-overview}

\ours\ is based on two key ideas.

\vspace{.3em}
\noindent
\textbf{Key idea 1: Atomic fact as a unit.}
Long-form text consists of many pieces of information that can each be either true or false.
Prior work has explored using a sentence as a unit; however, even a single sentence is a mix of supported and unsupported facts, e.g., in 40\% of the cases with ChatGPT.
Previous and concurrent work either (1) defines an additional label of \texttt{partial support}~\citep{manakul2023selfcheckgpt,liu2023evaluating} whose definition may be subjective and can lead to low agreement, 
or (2) takes the strictest definition of \texttt{support} that requires every piece of information to be supported~\citep{rashkin2021measuring,gao2022attributed}, which ignores the partial support cases, e.g., assigning 0.0 to both generations in Figure~\ref{fig:teaser} even though the first generation is considerably more accurate than the second.
\myskip{
\begin{table}[t]
    \footnotesize \setlength{\tabcolsep}{0em}
    \centering
    \begin{tabular}{p{1.2cm}p{6cm}}
        \toprule
        \multicolumn{2}{l}{
        \textbf{Sentence:} Thierry Henry (born 17 August 1977) is a French} \\
        \multicolumn{2}{l}{professional football coach, pundit, and former player.} \\
        \midrule
        \textbf{Fact 1:} &    
        Thierry Henry was born on 17 August 1977. \\
        \textbf{Fact 2:} &    
        Thierry Henry is French. \\
        \textbf{Fact 3:} &
        Thierry Henry is a professional football coach. \\
        \textbf{Fact 4:} &
        Thierry Henry is a football pundit.  \\
        \textbf{Fact 5:} &
        Thierry Henry is a former football player. \\
        \bottomrule
    \end{tabular}
    \caption{
      Examples of our process of decomposing sentences into \emph{atomic facts}, each conveying a single piece of information.
    }\label{tab:example-atomic-fact-decomposition}
\end{table}}

In this paper, we define an atomic fact as a short sentence conveying one piece of information (examples in Figure~\ref{fig:teaser}), similar to summarization content units~\citep{nenkova2004evaluating}.
An atomic fact is a more fundamental unit than a sentence for a piece of information and provides a more fine-grained evaluation, e.g., in Figure~\ref{fig:teaser}, rating the first generation higher than the second.

\vspace{.3em}
\noindent
\textbf{Key Idea 2: Factual precision as a function of a given knowledge source.}
Prior work often considers factual precision as a single global truth~\citep{manakul2023selfcheckgpt}.
In contrast, we adopt a perspective that the truthfulness of a statement should depend on a particular knowledge source that end users consider to be trustworthy and reliable.
Therefore, instead of whether an atomic fact is globally true or false, we consider whether it is {\em supported} by a given source of knowledge.
This has been used in the fact verification literature~\citep{wadden2022scifact} where conflict of information between different sources is relatively common.

\vspace{.3em}
\noindent
\textbf{Definition.}
Let $\mathcal{M}$ be a language model to be evaluated,
$\mathcal{X}$ be a set of prompts,
and $\mathcal{C}$ be a knowledge source.
Consider a response $y = \mathcal{M}_x$ for $x \in \mathcal{X}$
and $\mathcal{A}_y$, a list of atomic facts in $y$.
A \ours\ of $\mathcal{M}$ is defined as follows.
\begin{align*}
    \begin{gathered}
        f(y) = \frac{1}{|\mathcal{A}_y|}\sum_{a \in \mathcal{A}_y}\mathbb{I}[a \text{~is supported by~} \mathcal{C}], \\
        \text{\ours}(\mathcal{M}) =
        \mathbb{E}_{x \in \mathcal{X}}[f(\mathcal{M}_x) | \mathcal{M}_x\text{~responds}].
    \end{gathered}
\end{align*} {\em $\mathcal{M}_x$ responds} means $\mathcal{M}$ did not abstain from responding to the prompt $x$.
\myskip{
\vspace{10em}
Let $x$ be a long-form generation from an LM that conveys a series of atomic facts, i.e., pieces of information, $x_1, x_2, \cdots, x_n$, and $\mathcal{C}$ be a knowledge source.
We define \textbf{\ours} as a function that maps $x$ and $\mathcal{C}$ into:
\begin{equation*}
    \text{\ours}(x, \mathcal{C}) = \frac{1}{n} \sum_{i=1}^n \mathbb{I}[x_i \text{~is supported by~} \mathcal{C}].   
\end{equation*}
\begin{equation*}
    \text{\ours}(\mathcal{M}, \mathcal{C}) = \frac{
        \sum_{x \in \mathcal{X}} f(x, \mathcal{C}) \mathrm{DoesRespond}(x)
    }{
        \sum_{x \in \mathcal{X}} \mathrm{DoesRespond}(x)
    }
\end{equation*}}
This definition assumes the following:
\vspace{-.3em}
\begin{enumerate}[leftmargin=15pt]\itemsep -.2em
    \item Whether or not an atomic fact is supported by $\mathcal{C}$ is undebatable.
    \item Every atomic fact in $A_y$ has an equal weight of importance, following \citet{krishna-etal-2023-longeval}. 
    \item Pieces of information in $\mathcal{C}$ do not conflict or overlap with each other.
\end{enumerate}
In the rest of the paper, we propose to use people biographies as $\mathcal{X}$ and Wikipedia as $\mathcal{C}$ because they satisfy these assumptions to a reasonable degree (Section~\ref{subsec:data}).
We discuss in which cases these assumptions hold or may not hold in more detail in the Limitation section. 

\ours\ considers {\em precision} but not {\em recall}, e.g., a model that abstains from answering too often or generates text with fewer facts may have a higher \ours, even if these are not desired.
We leave the evaluation of factual recall for future work (more discussion in the Limitation section). 
\subsection{Studied LMs}
We evaluate three LMs (referred to as \subjectLM{}, 
an LM as a subject):
(1) \textbf{InstructGPT} (\texttt{text-davinci-003}, updated from \citet{ouyang2022training}),
(2) \textbf{ChatGPT}~\citep{chatgpt}, 
and (3) \textbf{PerplexityAI},\footnoteref{fn:ppl} which incorporates a search engine with a language model. 

\subsection{Data}
\label{subsec:data}

We perform human evaluation of factual precision based on our definition. We prompt the \subjectLM{} to generate {\em people biographies} and evaluate them against Wikipedia for the following reasons.
\vspace{-.3em}
\begin{itemize}[leftmargin=15pt]\itemsep -.2em
    \item Biographies are objective (not subjective or debatable) and contain specific (not vague) information, satisfying Assumption 1 in Section~\ref{subsec:data-overview}. 
    \item Biographies allow evaluation across diverse nationalities, professions, and levels of rarities.
    \item Wikipedia offers reasonable coverage of information about people and is reasonably self-consistent,\footnote{See Appendix~\ref{subsec:annotation-qualitative} for a related analysis.} satisfying Assumption 3. 
\end{itemize}

\paragraph{Data collection.}
We carefully design an annotation pipeline to assign a factual precision to a long-form generation through the following steps.

\vspace{.4em}
\noindent \textbf{Step 0: Sampling people entities.}
We sample 183 people entities from Wikidata who have 
corresponding Wikipedia pages.
We sample entities to annotate from a uniform distribution over categories defined in Appendix~\ref{app:data-category}. 

\vspace{.4em}
\noindent \textbf{Step 1: Obtaining generations.} We feed a prompt \texttt{``Tell me a bio of <entity>''} to the \subjectLM{} and take a generation as it is. 
We implement rules to identify generations that abstain from answering and filter them out. 

\vspace{.4em}
\noindent \textbf{Step 2: Atomic facts generation.}
Human annotators break a generation into a series of atomic facts.
To save annotation time, we provide atomic facts broken down by InstructGPT which human annotators can take and revise. Details in Appendix~\ref{app:atomic-fact-generation}.

\vspace{.4em}
\noindent \textbf{Step 3: Labeling factual precision \& editing.}
We ask another set of human annotators to assign each atomic fact one of three labels. If the atomic fact is clearly not related to the prompt, and thus should be removed from the bio without a validation step, they assign \irLabel. If the fact is relevant, they validate the fact based on the English Wikipedia, and label either \sLabel\ or \nsLabel. 

\vspace{.3em}
We recruit freelancers through Upwork and pay 15--25 USD per hour. Annotation requires extensive effort and time, leading to the cost of \$4 per generation. 
We assign two freelancers for the 10\% of the data and calculate the agreement rate: 96\%, 90\% and 88\% for InstructGPT, ChatGPT and PerplexityAI, respectively. More details are provided in Appendix~\ref{app:annotator-recruitment}.

\begin{table}[t]
    \footnotesize \centering
    \begin{tabular}{l @{\hspace{0.5em}} r @{\hspace{1em}} r @{\hspace{1em}} r}
        \toprule
            & InstGPT & ChatGPT & PPLAI \\
        \midrule
            Use search & \xmark & \xmark & \cmark \\
            \% responding & 99.5 & 85.8 & 90.7 \\
            \# tokens / response & 110.6 & 154.5 & 151.0 \\
            \# sentences / response & 6.2 & 7.9 & 9.8 \\
            \# facts / response & 26.3 & 34.7 & 40.8 \\
        \midrule
            \multicolumn{4}{c}{\em Statistics of the labels} \\
                \sLabel & 42.3 & 50.0 & 64.9 \\
                \nsLabel           & 43.2 & 27.5 & 11.1  \\
            \irLabel              
                & 14.0 & 8.3 & 14.8 \\
            Abstains from answering & 0.5 & 14.2  & 9.3 \\
        \midrule
            \textbf{\ours} & \textbf{42.5} & \textbf{58.3} & \textbf{71.5} \\
        \bottomrule
    \end{tabular}
    \caption{
      Statistics of the data and \ours{} results.
      InstGPT and PPLAI respectively refer to InstructGPT and PerplexityAI.
      {\em \% responding} indicates \% of generations that do not abstain from responding.
      {\em \# tokens} is based on white space. 
    }\label{tab:longform-data-statistics}
\end{table}
\myskip{
\begin{figure}[t]
\centering \footnotesize
\resizebox{1.7\columnwidth}{!}{
    \includegraphics[trim={5.3cm 8cm 6cm 1.7cm},clip]{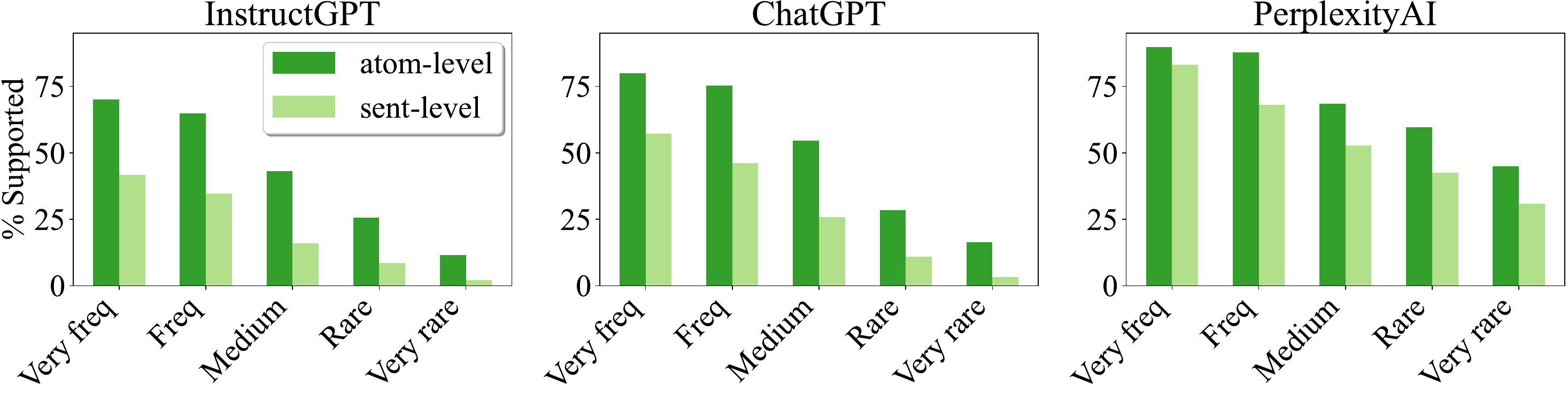}%
}
\resizebox{1.7\columnwidth}{!}{
    \includegraphics[trim={5.3cm 8cm 6cm 2.7cm},clip]{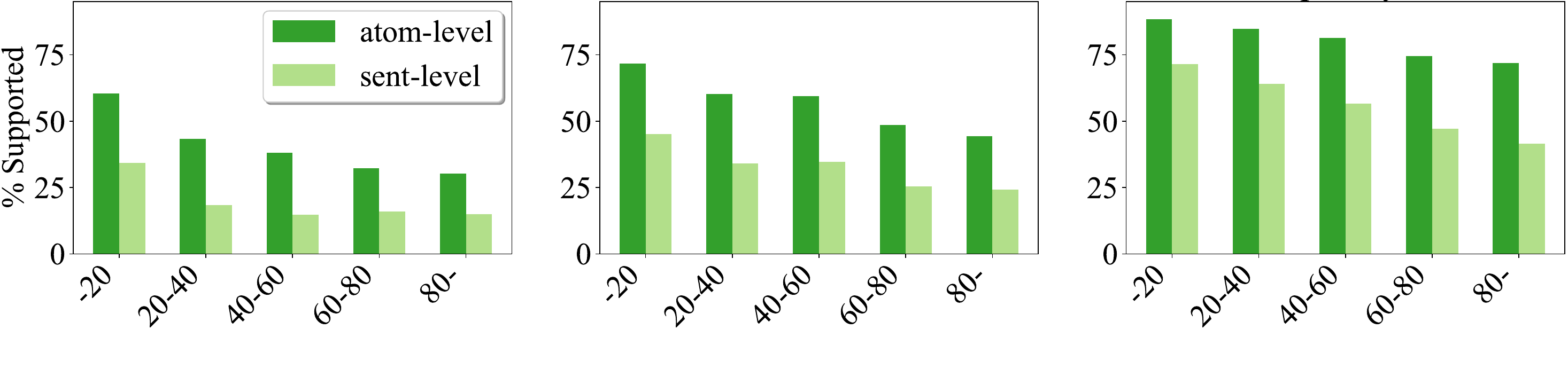}%
}\vspace{-1.5em}
\caption{
    \textbf{Top:} \% of {\em Supported} across varying frequency levels of human entities.
    \textbf{Bottom:} \% of {\em Supported} over relative positions in generation.
    There are significantly fewer supported facts (more precision errors) as the rarity of the entities increases and the position of the fact is later.
}\label{fig:benchmark-ablation}
\end{figure}
}

\subsection{Results}\label{subsec:annotation-results}
Statistics of the data and results are reported in Table~\ref{tab:longform-data-statistics}.

\vspace{-.3em}
\paragraph{All \subjectLM{}s struggle with factual precision errors.}
InstructGPT and ChatGPT achieve \ours{}s of 42.5\% and 58.3\%, respectively.
PerplexityAI, which uses a commercial search engine and thus should have a perfect \ours\ if directly copying 
the text from the correct Wikipedia page, attains a \ours\ of 71.5\%.
We provide a qualitative analysis of its error cases in the last paragraph of this section.

ChatGPT and PerplexityAI often abstain from answering which presumably improves their factual precision.
InstructGPT rarely abstains from answering, likely because it is not trained to do so.

Irrelevant facts either
(a) have dependencies on previous facts in a generation that turn out to be unsupported, 
or (b) are irrelevant to the prompt independent from other facts in a generation (examples in Appendix~\ref{app:example-annotated-data}).
We find that (b) rarely happens with InstructGPT and ChatGPT but happens considerably with PerplexityAI, because PerplexityAI often directly copies search results even if they are largely irrelevant to the input prompt.
This is in agreement with a concurrent work from \citet{liu2023evaluating} that shows generative search engines like PerplexityAI copy incorrect search results and generate text that is irrelevant to the input query.


\myskip{
\vspace{-.3em}
\paragraph{Irrelevant facts are frequent in PerplexityAI.}
Irrelevant facts are those that either
(a) have dependencies to previous facts in a generation that turn out to be unsupported, 
or (b) are irrelevant to the prompt independent from other facts in a generation (examples in Appendix~\ref{app:example-annotated-data}).
We find that (b) rarely happens with InstructGPT and ChatGPT, but happens considerably with PerplexityAI, because PerplexityAI often directly copies search results even if they are largely irrelevant to the input prompt.
This is in agreement with a concurrent work from \citet{liu2023evaluating} that shows generative search engines like PerplexityAI copy incorrect search results and generate text that is irrelevant to the input query.
}

\begin{table*}[t]
    \center  \myfontsize 
    \setlength{\tabcolsep}{0.5em}
    \begin{tabular}{p{2.3cm}p{0.4cm}p{12.5cm}}
        \toprule
            Category & \% & Example \\
        \midrule
            Single-sentence contradiction & 33.3 &
            \myblue{Gen} On November 25th, 2023, Glover Teixeira became an American citizen. \mypink{Wiki} In November 2020, Teixeira became an American citizen.\\
            (words) && \myblue{Gen} [Eric Hacker] was named the International League Pitcher of the Year. \mypink{Wiki} [Eric Hacker] was named the IL Pitcher of the Week. \\
        \cmidrule(lr){1-3}
            Single-sentence contradiction  & 10.0 &
            \myblue{Gen} William Waldegrave's grandfather was James II and VII. \mypink{Wiki} His father's title was created ... for the diplomat and ambassador James Waldegrave, 1st Earl Waldegrave, whose grandfather was James II and VII.
            \\
            (beyond words) && \myblue{Gen} She has appeared in several successful films such as (...) and Zero (2018). \mypink{Wiki}: Zero was a commercial failure. \\
        \cmidrule(lr){1-3}
            Page-level contradiction & 23.3 & \myblue{Gen} Some of [Julia Faye's] notable films include ... "Cleopatra" (1934). \mypurple{Comment} No mention of {\em Cleopatra} on the {\em Julia Faye} page, and no mention of {\em Julia Faye} on the {\em Cleopatra} page. \\
            && \myblue{Gen} [Kang Ji-hwan] has donated money to various charities and organizations over the years. \mypurple{Comment} No such mention on the {\em Kang Ji-hwan} page.
            \\
        \cmidrule(lr){1-3}
            Subjective & 16.7 & \myblue{Gen} His achievements, as an actor and as a cultural force, will surely prove to be as heroic as those of the characters he portrayed. \mypink{Wiki} Culture writer Steve Rose, in The Guardian, wrote:
            ``Chadwick Boseman began his career playing African American icons and pioneers; he ends it as one himself. His [...] achievements, as an actor and as a cultural force, will surely prove to be as heroic as those of the characters he portrayed.''\\
        \cmidrule(lr){1-3}
            Fact is irrelevant & 3.3 & \myblue{Gen} [Zamfir Arbore]'s life is not well-documented, and there is little information available about him.\\
        \cmidrule(lr){1-3}
            Wiki is inconsistent \& wrong & 3.3 & \myblue{Gen} Kick (2014) that brought [Sajid Nadiadwala] various debutant director awards. \mypink{Wiki} 2015, IIFA Award for Debut Director, Kick. (...) Kick brought him various debutant director awards. \mypurple{Comment} The first text is from a table that indicates he won one award (accurate). The second is inaccurate, incorrectly citing a news article. \\
        \cmidrule(lr){1-3}
            Annotation error & 10.0 &  \myblue{Gen} [Zamfir Arbore] was part of the staff of Românul. \mypink{Wiki} The Românul staff came to include Zamfir Arbore. \mypurple{Comment} Mentioned in the {\em Românul} page but not in the {\em Zamfir Arbore} page. \\
        \bottomrule
    \end{tabular}
    \caption{Categorization of precision errors (\nsLabel) from PerplexityAI (Section~\ref{subsec:annotation-qualitative}).
    \myblue{Gen} indicates the generation from PerplexityAI, and \mypink{Wiki} indicates evidence text from Wikipedia. \mypurple{Comment} indicates our comments.
    }\label{tab:ppl-error-analysis}
\end{table*}

\begin{figure}[t]
\centering \footnotesize
\resizebox{\columnwidth}{!}{
    \includegraphics[trim={22.5cm 8cm 6cm 1.7cm},clip]{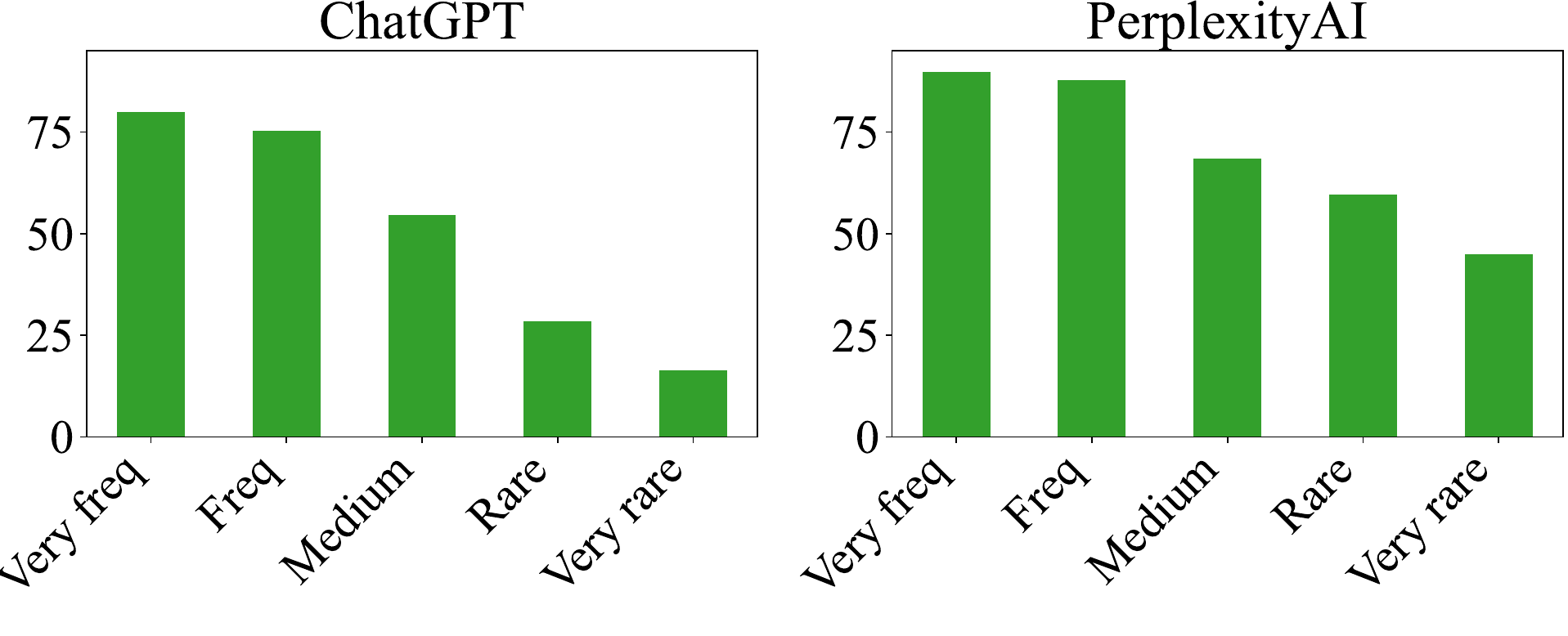}%
}
\resizebox{\columnwidth}{!}{
    \includegraphics[trim={22.5cm 8cm 6cm 2.7cm},clip]{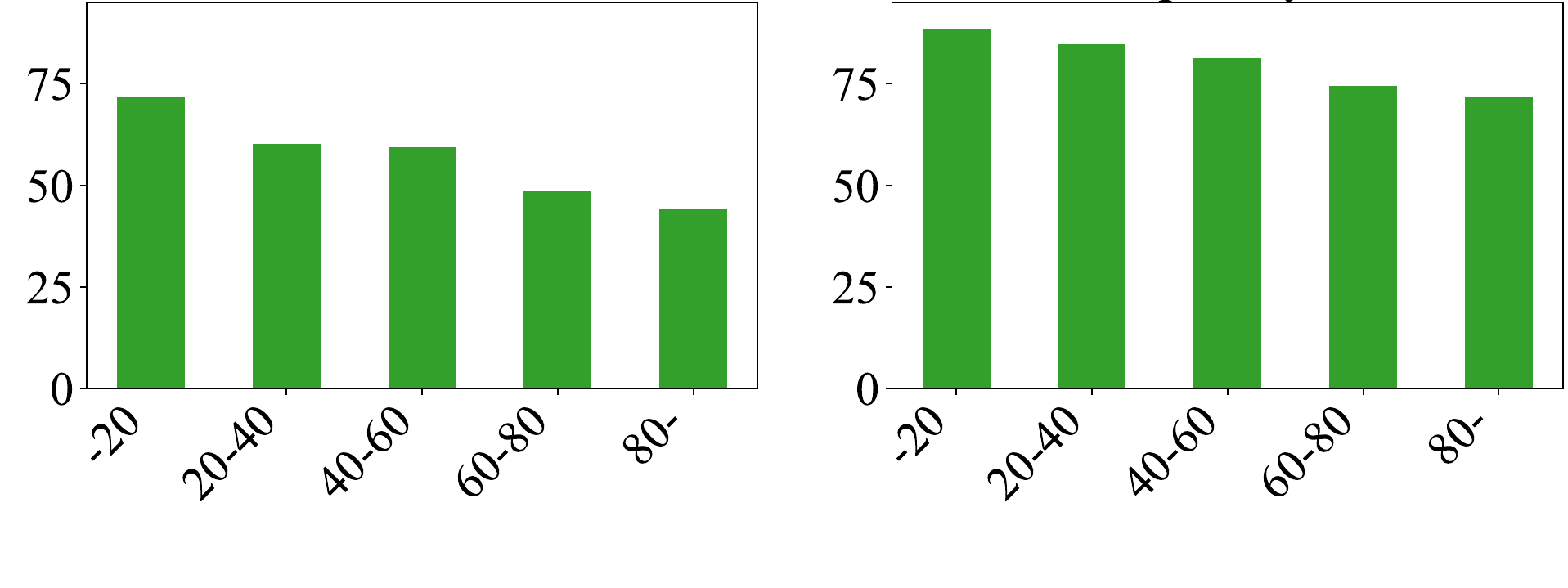}%
}\vspace{-1.5em}
\caption{
    \ours\ across varying frequency levels of human entities (\textbf{top}) and relative positions in a generation (\textbf{bottom}). \ourScores\ are lower as the rarity of the entities increases and the position of the fact is later.
}\label{fig:benchmark-ablation}
\end{figure}

\vspace{-.3em}
\paragraph{Error rates are higher for rarer entities.} 
Figure~\ref{fig:benchmark-ablation} (top) shows factual precision over varying frequency levels of topic entities (humans) in the pretraining corpora (see Appendix~\ref{app:data-category}).
There is a notable decrease in \ours\ as the rarity of entities increases, consistently across all \subjectLM{}s.
This is in agreement with \citet{kandpal2022large} and \citet{mallen2022not} which show that short question answering (QA) accuracy is highly correlated with the entity frequencies in the pretraining data. 
However, in contrast to \citet{kandpal2022large} and \citet{mallen2022not} who report QA accuracy of models with retrieval 
is robust to the rarity of entities,
\ours\ of PerplexityAI still significantly drops as entities are rarer: a relative drop of 50\% and 64\% observed at the atomic-level and sentence-level, respectively.

\vspace{-.3em}
\paragraph{Error rates are higher for facts mentioned later in the generation.} 
Figure~\ref{fig:benchmark-ablation} (bottom) reports factual precision over relative positions in a generation. 
Across all LMs, the later part of the generation has significantly worse precision. This is likely because (a) information mentioned earlier is more frequently mentioned in the pretraining data (e.g., nationality, profession), and (b) error propagation affects the later part of the generation.
This also implies that evaluating LMs solely based on short answers may not provide an adequate assessment of their factual precision, as it fails to account for errors that arise in the later stages of generation.

\vspace{-.3em}
\paragraph{Qualitative analysis of \nsLabel.}
One of the surprising findings in our empricial analysis is that a \ours\ of PerplexityAI (71.5\%) is lower than expected despite having access to the search engine. To better understand its errors, we categorize 30 random samples whose label is \nsLabel\ (Table~\ref{tab:ppl-error-analysis}).
\vspace{-.3em}
\begin{itemize}[leftmargin=12pt]\itemsep -.2em
    \item Single-sentence contradiction: A single sentence from Wikipedia provides direct contradiction to the generation, either at a word level (numbers, dates, or entities) or beyond. 
    \item Page-level contradiction: Errors found after reading the entire page, often because a fact that should have been mentioned in Wikipedia if true is missing, e.g., whether the subject appears in a particular film.
    \item Subjective: Generation is subjective, often because PerplexityAI copies subjective text from Wikipedia, e.g., directly copying a quote from a journalist without realizing it.
    \item Fact is irrelevant: Generation is irrelevant to the subject due to a search error.
    \item Wiki is inconsistent \& wrong: In the example, Wikipedia indicates that the subject won one award from the film Kick, but also includes text that they won multiple awards from Kick, which is inaccurate and cited a news article that does not support the claim.
    \item Annotation error: Annotators assign incorrect labels, typically because the information is not mentioned in the subject's Wikipedia page (likely because it is insignificant). 
\end{itemize}

\noindent
We also find that, although PerplexityAI provides citations to the references, citations have little correlation with factual precision. 36.0\% and 37.6\% of supported and unsupported sentences have citations, respectively. 
Together with independent findings from \citet{liu2023evaluating}, 
this indicates that commercial LMs that incorporate search and provide citations may not be as reliable as expected.

\vspace{.2em}
More analysis is provided in Appendix~\ref{subsec:annotation-qualitative}. 

\section{Estimating \ours\ for Automatic Evaluation}\label{sec:method}
\newcommand{\myO}{$+$}
\newcommand{\myU}{$-$}
\newcommand{\strongO}{\myO}
\newcommand{\strongU}{\myU}

\begin{table*}[t]
    \center
    \footnotesize
    \begin{tabular}{ll @{\hspace{0.2em}}
            R{0.6cm}
            R{1cm} @{\hspace{0.1em}} R{1cm} @{\hspace{0em}} R{0.3cm}
            R{1cm} @{\hspace{0.1em}} R{1cm} @{\hspace{0em}} R{0.3cm}
            R{1cm} @{\hspace{0.1em}} R{1cm} @{\hspace{0em}} R{0.3cm}
            @{\hspace{-0.3em}} 
            R{1.2cm}
        }
        \toprule
            & \multirow{2}{*}{Evaluator} & \multirow{2}{*}{retrv} & \multicolumn{3}{c}{\textsc{subj}: InstGPT} & \multicolumn{3}{c}{\textsc{subj}: ChatGPT} & \multicolumn{3}{c}{\textsc{subj}: PPLAI} & \multirow{2}{*}{ranking} \\
            \cmidrule(lr){4-6} \cmidrule(lr){7-9} \cmidrule(lr){10-12}
            & & & ER & FS && ER & FS && ER & FS && \\
        \midrule    
            & Human &                     && 42.5 &&& 58.3 &&& 71.5 \\
        \midrule
            \parbox[t]{2mm}{\multirow{3}{*}{\rotatebox[origin=c]{90}{\myfontsize  Trivial}}} &
            Always \sLabel &            & 57.5 &100.0 & \strongO & 41.7& 100.0 &\strongO& 28.5 &100.0 &\strongO & \xmark \\
            & Always \nsLabel &           & 42.5 &0.0 &\strongU& 58.3 &0.0 & \strongU & 71.5& 0.0 &\strongU & \xmark \\
            & Always Random &             & 7.5 &50.0 & \myO & 8.3 &50.0 &\myU& 21.5 &50.0 &\strongU & \xmark \\
        \midrule
            \parbox[t]{2mm}{\multirow{4}{*}{\rotatebox[origin=c]{90}{\myfontsize I-LLAMA}}} &
            No-context LM & \xmark      & 7.1 &49.6 & \myO & 7.8 &50.5 &\myU& 34.7 &36.8 & \strongU & \xmark  \\
            & NP & \cmark                & 14.8 &57.3 & \myO & 13.7& 72.0 &\myO& 1.4 &72.9 & & \gmark \\
            & \rtg & \cmark               & 14.1 &56.6 & \myO & 17.1& 75.4 &\myO& \best{0.1} & 71.6 & & \xmark \\
            & \rtg\ + NP & \cmark        & \best{1.4} &41.1 & & \best{0.4} &58.7 && 9.9 &61.6 &\myU & \gmark \\
        \midrule
            \parbox[t]{2mm}{\multirow{3}{*}{\rotatebox[origin=c]{90}{\myfontsize  ChatGPT}}} &
            No-context LM & \xmark      & 39.6& 82.1 & \strongO & 31.7& 90.1& \strongO& 3.3 &74.8 & & \xmark \\
            & \rtg & \cmark               & 5.1 &47.6 & \myO & 6.8& 65.1& \myO & 0.8& 72.3 & & \gmark \\
            & \rtg\ + NP & \cmark        & 5.2 &37.3 &\myU& 4.7& 53.6 && 8.7 &62.8 &\myU & \gmark \\
        \bottomrule
    \end{tabular}
    \caption{
        Results on \textbf{Error Rate (ER)} along with \ours{}s estimated by each model (\textbf{FS}).
        `{\em retrv}' indicates whether or not retrieval is being used, and
        `{\em ranking}' \gmark\ indicates whether the ranking between three \subjectLM{}s rated by the model is consistent to the ground truth ranking.
        \myO\ and \myU\ respectively indicate the estimation is an overestimation and an underestimation by more than 5\% in absolute.
        \best{Red Bold} indicates the best (lowest) ER.
        See Appendix~\ref{app:fone} for the results in other metrics that consider individual judgments instead of aggregated ones.
    }\label{tab:validation-error-rate}
\end{table*}

Human evaluation of factual precision is costly (\$4 per generation)~\citep{bohnet2022attributed,krishna-etal-2023-longeval}
because validating every atomic fact against a large knowledge source is time-consuming, and one generation contains many (26--41) atomic facts. This prevents LM developers and practitioners from evaluating the factual precision in long-form generation of a new \subjectLM{} at scale.
In this context, we introduce a model that \textbf{estimates} \ours. This estimator takes a set of generations and automatically computes a \ours, and can be applied to any \subjectLM.

We describe our model (Section~\ref{subsec:models-validation}) and demonstrate its accuracy against human evaluation (Section~\ref{subsec:models-validation-result}). 
\ours\ estimated by our model is then used to evaluate twelve LMs (Section~\ref{sec:eval-new-lms}).


\subsection{Model}\label{subsec:models-validation}

Our estimator of \ours\ first breaks a generation into a series of atomic facts and then validates each against the given knowledge source. 
We find taking atomic facts generated by InstructGPT (used in data collection in Section~\ref{subsec:data}) effective and close to human, consistent with findings from prior work~\citep{chen2022generating}.
\pw{are there details of how this was measured? add a ref?}
This section thus focuses on how to validate each atomic fact against a given knowledge source.

The validation is based on zero-shot prompting of an LM referred to as an \textbf{\evalLM} to distinguish from an \subjectLM.
Specifically, a {\em prompt}---whose construction methods differ across four variants---is fed into an \evalLM.
The prediction is then made by comparing the conditional probability of \texttt{True} and \texttt{False} from the \evalLM.
If the logit values are unavailable (e.g., commercial LMs like ChatGPT), the prediction is made based on whether the generated text contains \texttt{True} or \texttt{False}.\footnote{
    In Appendix~\ref{app:val-abl}, we compare with an alternative prompting that generates a question and compares the answer to it and the expected answer~\citep{kryscinski2019evaluating,wang2020asking,gao2022attributed,manakul2023selfcheckgpt}.
    We empirically find that our prompting performs better due to the lack of control over the questions being generated.
}

\noindent 
The four variants we consider are as follows.

\vspace{.3em}
\noindent 
\textbf{No-context LM} uses \texttt{<atomic-fact> True or False?} as a prompt, closely resembling \citet{kadavath2022language}.\footnote{In Appendix~\ref{app:val-abl}, we also compare with Self-check LM, a concurrent work from \citet{manakul2023selfcheckgpt}. We do not include it in the main paper because it has strong restrictions, e.g., requires access to the \subjectLM\ at evaluation time and cannot be applied to PerplexityAI with nondeterministic outputs. 
}

\vspace{.3em}
\noindent \textbf{\rtg} retrieves passages from the given knowledge source and then prompts the \evalLM. It first retrieves $k$ passages, constructs the prompt by concatenating retrieved passages, the given atomic fact, and \texttt{``True or False?''}, and feeds it to the \evalLM\ to get the prediction.

\vspace{.3em}
\noindent \textbf{Nonparametric Probability (NP)} makes a judgment based on a nonparametric likelihood. 
It masks out each token in the atomic fact, computes its likelihood using a nonparametric masked LM~\citep{min2022nonparametric}, averages probabilities over all tokens, and makes a prediction based on thresholding.

\vspace{.3em}
\noindent \textbf{\rtg\ + NP} is an ensemble of \rtg\ and NP which assigns \sLabel\ only if both methods assign \sLabel. 

\vspace{.3em}
We use LLAMA 7B trained on Super Natural Instructions (Inst-LLAMA, \citealp{touvron2023llama,wang2022super}) and ChatGPT as an \evalLM, and Generalizable T5-based Retrievers (GTR, \citet{ni2021large}) for passage retrieval.
See Appendix~\ref{subsubsec:setup-validation} for more implementation details.

\subsection{Evaluation of Estimators}\label{subsec:models-validation-result}

\paragraph{Metrics.}
We report \textbf{Error Rate (ER)}---the difference between the ground truth and the estimated \ours---as well as whether the estimated \ourScores\ preserve the ranking between three \subjectLM{}s.
Appendix~\ref{app:fone} discusses results with other metrics that consider individual judgments instead of aggregated judgments.
We use the data in Section~\ref{subsec:data} as evaluation data.

\vspace{.3em}
\noindent
Results are reported in Table~\ref{tab:validation-error-rate}.

\vspace{-.3em}
\paragraph{Retrieval significantly helps.}
Models that use retrieval are consistently better than No-context LM which either has a significantly high ER or does not preserve ranking between three \subjectLM{}s.
This is likely because the \evalLM\ has not memorized every factual information about the topic entity, thus benefiting from retrieval providing factual context.
%
Nonetheless, just using \rtg\ may overestimate \ours, e.g., by up to 17\% with Inst-LLAMA, when a \subjectLM{} is InstructGPT or ChatGPT.
In this case, ensembling \rtg\ and NP reduces an error rate by a significant margin. 
When a \subjectLM{} is PerplexityAI, single methods (either \rtg\ or NP) give a low ER, and ensemble methods have a higher ER due to an underestimation of \ours.

\vspace{-.3em}
\paragraph{ChatGPT is not always the best.}
Our results show that ChatGPT is not necessarily better than Inst-LLAMA. We investigate this further in Appendix~\ref{app:val-abl}. In summary, ChatGPT is better at validating each individual atomic fact. However, most errors from ChatGPT are incorrectly assigning \sLabel\ to unsupported facts, overestimating \ours.
In contrast, LLAMA+NP is not biased toward overestimation or underestimation of the factual precision, resulting in an aggregated factual precision to be closer to the ground truth.
This is similar to the trade-off between system-level and segment-level correlations in summarization evaluation, which often produce different rankings~\citep{bhandari-etal-2020-evaluating,deutsch2021towards}.

\vspace{-.3em}
\paragraph{The best estimator depends on the \subjectLM.}
While using retrieval is consistently better than No-context LM, the best variant of estimator depends on a \subjectLM: LLAMA+NP for InstructGPT and ChatGPT, and ChatGPT for PerplexityAI. 
Nevertheless, both evaluators give consistently correct ranking between three \subjectLM{}s, and Section~\ref{sec:eval-new-lms} show scores from two estimators are largely correlated across 10+ \subjectLM{}s (0.99 Pearson's $r$). We recommend users try both variants of our estimator when evaluating a new \subjectLM\ and report their correlation. 


\subsection{Evaluation of New LMs}\label{sec:eval-new-lms}Our estimator allows evaluating factual precision of a large set of new LMs at scale with no human efforts.
As a case study, we evaluate ten new LMs that came out within two months at the time of conducting experiments (Table~\ref{tab:new-lms}).
These LMs were evaluated on many benchmarks but not in factual precision of long-form generation since such evaluation is costly.
We aim to provide new insights on these LMs by estimating \ours\ of their long-form generations.

\subsubsection{Setup}
\begin{table}[t]
    \center \mysmallfontsize 
    \setlength{\tabcolsep}{0.25em}
    \begin{tabular}{l @{\hspace{0em}} c @{\hspace{0.8em}} c @{\hspace{0.2em}} c @{\hspace{0em}} r 
        }
        \toprule        
           {\subjectLM} & Base LM & Use other LMs & Open & Release \\
        \midrule
            InstructGPT & ? & ? & \xmark & Nov 2022  \\
            ChatGPT & ? & ? & \xmark & Nov 2022      \\
            GPT-4 & ? & ? & \xmark & Mar 2023        \\
            Alpaca \{7B,13B,65B\} & LLAMA & InstructGPT  & \cmark & Mar 2023 \\
            Vicuna \{7B,13B\} & LLAMA & ChatGPT & \cmark  & Mar 2023     \\
            Dolly 12B & Pythia 12B & N/A & \cmark & Mar 2023             \\
            Oasst-pythia 12B & Pythia 12B & N/A & \cmark & Mar 2023      \\
            StableLM-tuned 7B & StableLM-base & ChatGPT, GPT-4 & \cmark & Apr 2023 \\
            MPT Chat 7B & MPT 7B & ChatGPT & \cmark & May 2023   \\
        \bottomrule
    \end{tabular}
    \caption{
        A set of twelve LMs evaluated in Section~\ref{sec:eval-new-lms}. 
        All models are tuned for instruction following or chat.
        {\em Use other LMs} indicates whether the model is trained on any data that includes outputs of another model.
        {\em Open} indicates model weights 
        are publicly available.
    }\label{tab:new-lms}
\end{table}

\begin{figure*}[t]
\centering \footnotesize
\resizebox{2\columnwidth}{!}{
    \includegraphics[trim={0.3cm 0 0 1.1cm},clip]{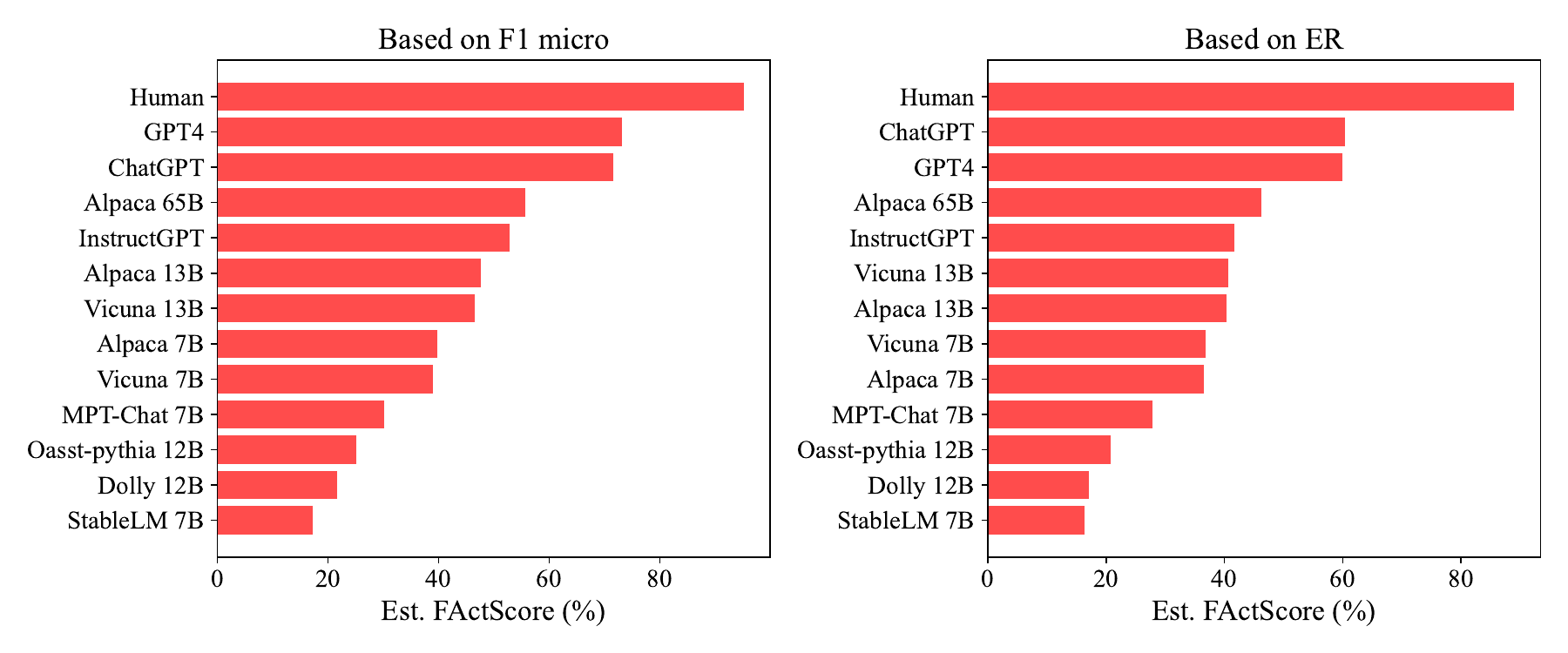}%
}\vspace{-1em}
\caption{
    Ranking between 13 subjects (human and 12 LMs), rated by the two best variants of our estimator: ChatGPT (\textbf{left}) and LLAMA+NP (\textbf{right}), both with retrieval.
    Scores from two metrics have a Pearson's $r$ of 0.99.
    See Table~\ref{tab:new-lms-statistics} for \% of responding and \# of atomic facts per response of each LM. 
    The variance in estimation based on different subsets of prompts is reported in Figure~\ref{fig:new-lms-ranking-error-bar} of Appendix~\ref{app:new-lms-additional}.
    }\label{fig:new-lms-ranking}
\end{figure*}

We evaluate 10 recently-released LMs as shown in Table~\ref{tab:new-lms}.
\textbf{GPT-4}~\citep{OpenAI2023GPT4TR} is a multimodal LM released by OpenAI available through an API.
\textbf{Alpaca}~\citep{alpaca} is based on LLAMA~\citep{touvron2023llama} fine-tuned on the instructions data based on InstructGPT following the recipe from~\citet{wang2022super}.
\textbf{Vicuna}~\citep{vicuna2023} is based on LLAMA fine-tuned on the outputs from ChatGPT available through ShareGPT.\footnote{
\href{https://sharegpt.com/}{\texttt{sharegpt.com}}
\textsuperscript{9}\href{https://huggingface.co/databricks/dolly-v2-12b}{\texttt{dolly-v2-12b}}
\textsuperscript{10}\href{https://www.databricks.com}{\texttt{databricks.com}}
\textsuperscript{11}\href{https://huggingface.co/OpenAssistant/oasst-sft-1-pythia-12b}{\texttt{oasst-sft-1-pythia-12b}}
\textsuperscript{12}\href{https://open-assistant.io/}{\texttt{open-assistant.io}}
\textsuperscript{13}\href{https://huggingface.co/stabilityai/stablelm-tuned-alpha-7b}{\texttt{StableLM-tuned-alpha-7b}}
\textsuperscript{14}\href{https://huggingface.co/stabilityai/stablelm-base-alpha-7b}{\texttt{stablelm-base-alpha-7b}}
\textsuperscript{15}\href{https://www.mosaicml.com/blog/mpt-7b}{\texttt{mosaicml.com/blog/mpt-7b}}
\textsuperscript{16}\href{https://huggingface.co/datasets/victor123/evol_instruct_70k}{\texttt{evol\_instruct\_70k}}}
\textbf{Dolly}\footnotemark[9] is Pythia 12B~\citep{biderman2023pythia} fine-tuned on DataBricks Dolly, human-written data created by Databricks.\footnotemark[10]
\textbf{Oasst-pythia}\footnotemark[11] is Pythia 12B fine-tined on human-written data collected through Open Assistant.\footnotemark[12]
\textbf{StableLM-tuned-alpha}\footnotemark[13] is based on StableLM-base-alpha\footnotemark[14] fine-tuned on the data used in the Alpaca data, DataBricks Dolly, the ShareGPT data, the GPT4All data~\citep{gpt4all} and Anthropic HH~\citep{bai2022training}.
\textbf{MPT Chat} is based on MPT 7B\footnotemark[15] fine-tuned on the ShareGPT data, the Alpaca data, Anthropic HH, HC3~\citep{guo-etal-2023-hc3}, and Evol-Instruct.\footnotemark[16]

\setcounter{footnote}{16} 

\begin{table}[t]
    \center
    \footnotesize
    \begin{tabular}{l @{\hspace{-0.8em}} r r
        }
        \toprule        
           \subjectLM & \% responding & \#facts / res \\
        \midrule
            GPT-4       & 88.2 & 60.8 \\
            Vicuna 13B & 76.6 & 50.9 \\
            Vicuna 7B & 91.0 & 45.6 \\
            Oasst-pythia 12B & 100.0 & 39.7 \\
            StableLM-tuned-alpha 7B &  66.6 & 38.0 \\
            MPT Chat 7B & 88.8 & 37.3 \\
            ChatGPT     & 84.2 & 37.0 \\
            InstructGPT & 99.8 & 27.7 \\
            Dolly 12B & 100.0 & 24.6 \\
            Alpaca 7B & 100.0 & 17.4 \\
            Alpaca 65B & 100.0 & 17.1 \\
            Alpaca 13B & 100.0 & 16.6 \\ 
        \midrule
            Human & 88.8 & 29.0 \\
        \bottomrule
    \end{tabular}
    \caption{
        Statistics of 500 model-generated bios in our unlabeled data from 12 LMs as well as human-written bios.
        {\em \% responding} indicates \% of generations that do not abstain from responding.
        {\em \#facts / res} indicates \# of atomic facts per response.
        LMs are sorted based on \# of facts per response.
        See Figure~\ref{fig:new-lms-ranking} for their \ourScores.
    }\label{tab:new-lms-statistics}
\end{table}

\vspace{.3em}
We prompt each \subjectLM\ to generate biographies of 500 human entities as done in Section~\ref{subsec:data} but with no overlap in entities.
We additionally include InstructGPT, ChatGPT, and human-written biographies obtained through DBPedia. 
Human-written biographies were unavailable for 11\% of entities which we consider as abstaining from responding.
See Table~\ref{tab:new-lms-statistics} for their statistics.
In total, we evaluate 6,500 generations from 13 subjects, which would have cost \$26K if they were evaluated by humans.

\subsubsection{Results}
Figure~\ref{fig:new-lms-ranking} shows the ranking between 13 subjects provided by the two best variants of our estimator whose scores are largely correlated, e.g., having a Pearson's $r$ of 0.99.
This evaluation allows a better understanding of these models, including:\vspace{-.3em}
\begin{itemize}[leftmargin=12pt]\itemsep -.2em
    \item All LMs are substantially less factual than humans. This is in contrast to prior work that claims LMs approach human performance, even for complex tasks~\citep{ding2022gpt,nori2023capabilities,lee2023benefits} even though the task of writing  biographies is fairly easy.
    \item GPT-4 and ChatGPT are comparable in factual precision. However, as reported in Table~\ref{tab:new-lms-statistics}, GPT-4 abstains from responding less (12\% vs. 16\%) and generates significantly more facts (61 vs. 37 per response).
    \item GPT-4 and ChatGPT are significantly more factual than public models.
    \item Within the same family of models that differ in sizes, there is a clear correlation between the model size and factual precision, e.g., Alpaca 65B > 13B > 7B, and Vicuna 13B > 7B.
    \item Alpaca and Vicuna achieve performance that is very close to each other within the same size of models, possibly because they share the same base model and similar training data.
    Nonetheless, as shown in Table~\ref{tab:new-lms-statistics}, 
    Vicuna generates significantly more atomic facts than Alpaca does (51 vs. 17 per response). Also, Alpaca never abstains from answering while Vicuna does. 
    \item Within public models, there are large gaps in factual precision even when the model size is similar, e.g., within the 7B models, Alpaca and Vicuna ($\sim40\%$) are more factual than MPT-Chat ($30\%$) and StableLM ($17\%$).
    Possible factors include the choice of the base LM, the data, and the training recipe~\citep{hoffmann2022training}. 
\end{itemize}

\myskip{
\begin{figure}[t]
\centering \footnotesize
\resizebox{1\columnwidth}{!}{
    \includegraphics[trim={20cm 0 0 1.1cm},clip]{figures/manylm_comparison.pdf}%
}\vspace{-.5em}
\caption{
    Ranking between 13 subjects (human and 12 LMs), rated by our estimator with the best ER.
    See Table~\ref{tab:new-lms-statistics} for \% of responding and \# of atomic facts per response of each LM. 
    The variance in estimation based on different subsets of prompts is reported in Figure~\ref{fig:new-lms-ranking-error-bar} of Appendix~\ref{app:new-lms-additional}.
    }\label{fig:new-lms-ranking}
\end{figure}}

\vspace{-.1em}
\noindent
We highlight that this evaluation only considers factual precision, specifically in people biographies. A holistic evaluation of LMs should include other aspects of generations such as fluency, coherence, relevance, consistency and creativity, which is out of scope of this paper.


\section{Conclusion and Future Work}\label{sec:concl}We introduced \ours, a new evaluation of the factual precision of long-form generation from LMs that breaks a generation down into a series of atomic facts and computes a fraction of facts supported by a given knowledge source.
We first performed extensive human evaluation, finding that commercial, state-the-art-art LMs---InstructGPT, ChatGPT, and search engine augmented, PerplexityAI---make a substantial amount of errors, e.g., having a \ours\ of 58\% in the case of ChatGPT.
Since human evaluation is time-consuming and costly, we proposed a model that estimates \ours, allowing an automatic evaluation of \ours.
We found our estimator based on retrieval over a knowledge source and competitive language models estimates \ours\ close to the ground truth, and showcased its application by evaluating 12 recently-released LMs that could have cost \$65K if evaluated by humans and providing insights about them.

Within four months since its initial release, \ours\ has actively been used in subsequent work, evaluating factual precision of recently-proposed models~\cite{ye2023flask,sun2023exploring,malaviya2023expertqa,dhuliawala2023chain}.
As future work, we suggest: (1) considering other aspects of factuality such as recall (coverage of factual information); (2) further improving the estimator for a better approximation of factual precision; and (3) leveraging \ours\ to correct model generations (briefly explored in Appendix \ref{sec:editing}).

\section*{Limitations}\label{sec:limitation}\paragraph{Scope of \ours.}
All of our experiments focus on people biographies and Wikipedia, because many LMs can generate biographies with objective and specific facts (rather than subjective and vague ones) and Wikipedia has a high coverage for them.
\ours\ can be applied to a broader domain, e.g., text about recent events whose knowledge source can be a collection of news articles, or text about scientific findings whose knowledge source can be a collection of scientific literature.
We present a proof of concept in Appendix~\ref{app:nlp-domain} and leave further study for future work.

Due to the assumptions made in Section~\ref{subsec:data-overview},
\ours\ is not applicable when the facts are more nuanced, open-ended, and debatable~\citep{chen2019seeing,xu2023critical} or with a knowledge source whose text frequently conflicts with each other~\citep{wadden2022scifact}.
Moreover, \ours\ may not be suitable for the human-written text that is nuanced and includes intentional or implicit deception.


\vspace{-.1em}
\paragraph{Limitation in our estimator.}
While our estimator closely approximates humans and provides consistent ranking over a large set of LMs, it is not perfect in individual judgments, and the best variant depends on the degree of how close a generation is to human-written text and its linguistic complexity.
Future work can investigate how the distribution of model generation affects the performance of the estimator and further improve the estimator.

\vspace{-.1em}
\paragraph{Beyond factual precision.}
\ours\ focuses on factual precision---whether each piece of information in a generation is factually supported by a reliable source of knowledge---which is only one aspect of the broader factuality problem.
For instance, \ours\ does not consider factual recall: the coverage of information in a generation.
\ours\ does not penalize a model that abstains from responding too frequently or generates fewer facts, which can be unfair since there is an inherent trade-off between precision and recall.
Moreover, the boundary between precision and recall is often blurry, e.g., it is possible that, even if every piece of information in a generation is supported, it misses a significant piece of information that should have been mentioned in order to be considered as correctly responding to the input prompt (example in Table~\ref{tab:recall}).
We leave a more holistic evaluation of factuality for future work, and recommend reporting \ours\ together with the \% of abstention and the average number of atomic facts (as we did in Section~\ref{sec:eval-new-lms}).

\begin{table}[t]
    \center \footnotesize
    \begin{tabular}{p{7.3cm}}
        \toprule        
           \textbf{Prompt:} Tell me a bio of Mary I of England. \\
           \textbf{Generation:}
           Mary I 
           was Queen of England and Ireland from July 1553 until her death. Mary was the daughter of King Henry VIII and his first wife, Catherine of Aragon. After her father annulled his marriage to Catherine, Mary’s status was in doubt, and she was excluded from the line of succession. Her half-brother, Edward VI, succeeded their father in 1547, and Mary had to live with the humiliation of being declared illegitimate. \\
        \bottomrule
    \end{tabular}
    \caption{An example whose factual precision is high but recall is low. The generation does not mention how Mary I of England got back to the line of succession and eventually became a queen. 
    \vspace{-.5em}
    }\label{tab:recall}
\end{table}

\section*{Acknowledgement}
We thank Yizhong Wang for sharing Instruction-tuned LLAMA and Alpaca models with varying sizes, and for sharing feedback on the \ours\ Python package. We thank experts in Upwork for annotating the data, and Dhruba Ghosh, Jiacheng Liu and Zeqiu Wu for participating in pilot annotation and sharing feedback. We thank Akari Asai, Yanai Elazar, UW NLP members, UMass NLP members, FAIR lab members for feedback and discussion on the paper.

This research was supported by
NSF IIS-2046248, NSF IIS-2202506, NSF IIS-2044660, ONR N00014-18-1-2826, ONR MURI N00014- 18-1-2670, DARPA under Contract No. FA8650-23-C-7316, an Allen Distinguished Award, and gifts from AI2.
The views, opinions and/or findings expressed are those of the author and should not be interpreted as representing the official views or policies of the Department of Defense or the U.S. Government.
Sewon Min is supported by a J.P. Morgan fellowship, and Kalpesh Krishna was supported by the Google PhD Fellowship.

\bibliography{anthology,custom}
\bibliographystyle{acl_natbib}

\clearpage
\appendix

\section{Details in Data Collection}\label{app:data-details}
\subsection{Sampling human entities}\label{app:data-category}

We sample 183 human entities to be annotated as follows. We first choose entities from Wikidata whose \texttt{instance of} is \texttt{human} and have corresponding Wikipedia pages.
We then categorize entities based on two dimensions: frequency and nationality, resulting in 20 categories. We then sample entities uniformly at random over all categories.

\vspace{-.2em}
\paragraph{Frequency.}
We compute {\texttt{freqValue}} as a maximum of the entity occurrence in Wikipedia provided by \citet{kandpal2022large} and the pageview count of the Wikipedia page following \citet{mallen2022not}. We found using one of them could lead to an underestimate of frequency levels due to failure in entity linking or mismatch in the Wikipedia page title, and taking a maximum of them provides a reasonable solution.
We then assign one of five categories: 
{`Very rare'} if {\texttt{freqValue}}$ \in [0, 10^2)$,
{`Rare'} if {\texttt{freqValue}}$ \in [10^2, 10^3)$,
{`Medium'} if {\texttt{freqValue}}$ \in [10^3, 10^4)$,
{`Frequent'} if {\texttt{freqValue}}$ \in [10^4, 10^5)$, and
{`Very frequent'} if {\texttt{freqValue}}$ \in [10^5,)$.

\vspace{-.2em}
\paragraph{Nationality.} We take \texttt{country of citizenship} from Wikidata and assign them one of four categories: `North America', `Europe \& Middle East', `Asia \& Pacific' and `Latin/South America \& Africa'.

\subsection{Details in generating atomic facts}\label{app:atomic-fact-generation}
We break out a generation automatically by splitting a generation into sentences, and feeding each sentence to InstructGPT (\texttt{text-davinci-003}) with a series of instructions to further break it down to a series of atomic facts.
The prompt to InstructGPT is provided in Table~\ref{tab:instruction-atomic-facts}.
Outputs from InstructGPT are used (1) to human experts for revision (Section~\ref{subsec:data}) and (2) for model-based evaluators (Section~\ref{sec:method}).
We find human experts split and merged atomic facts from InstructGPT for 18\% and 34\% of the cases, respectively.

\subsection{More details on annotator recruitment}\label{app:annotator-recruitment}
\begin{figure*}[t]
\centering \footnotesize
\resizebox{2\columnwidth}{!}{
    \includegraphics[trim={0 0 0 0},clip]{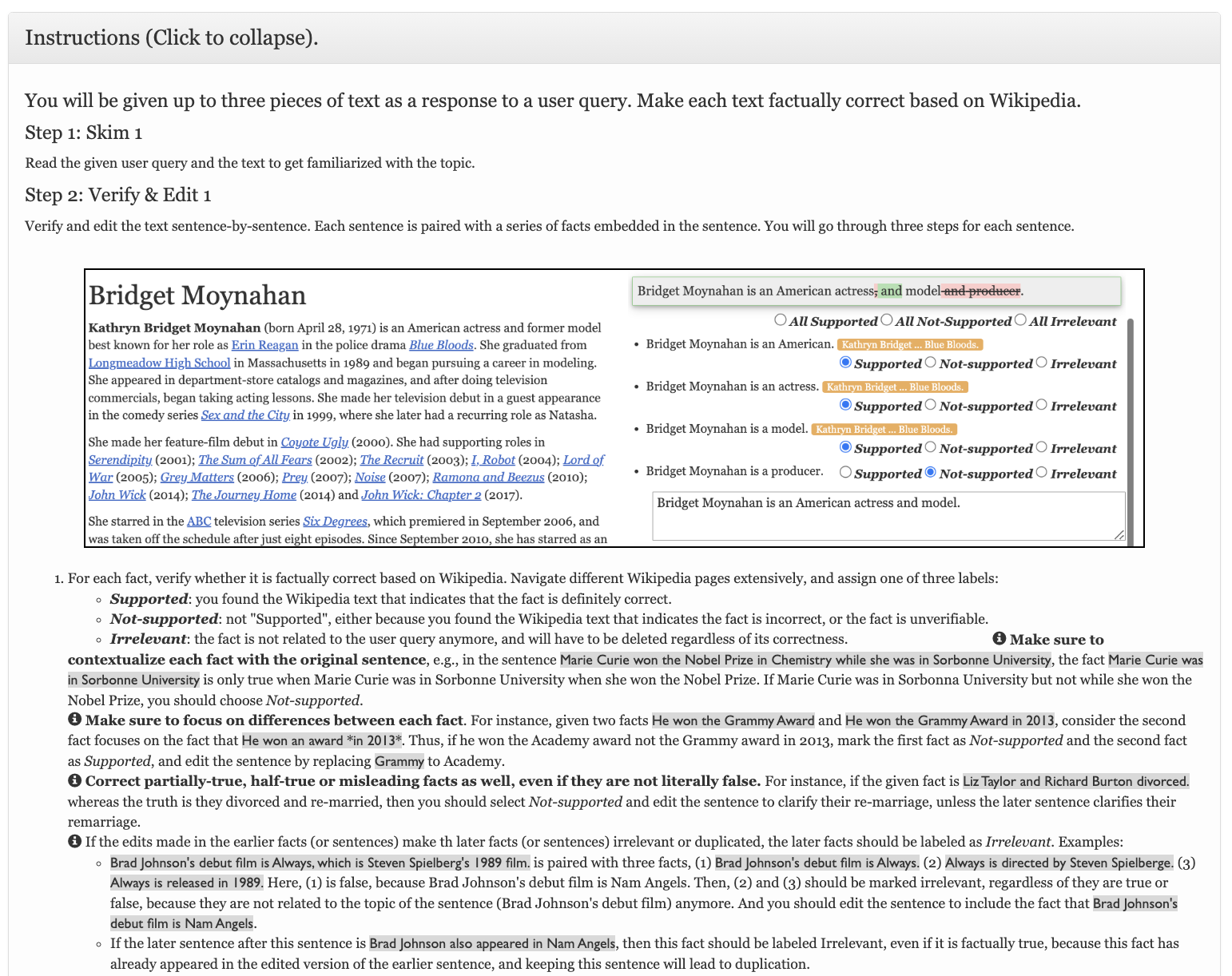}%
}
\resizebox{1.98\columnwidth}{!}{
    \includegraphics[trim={0 0 0 0},clip]{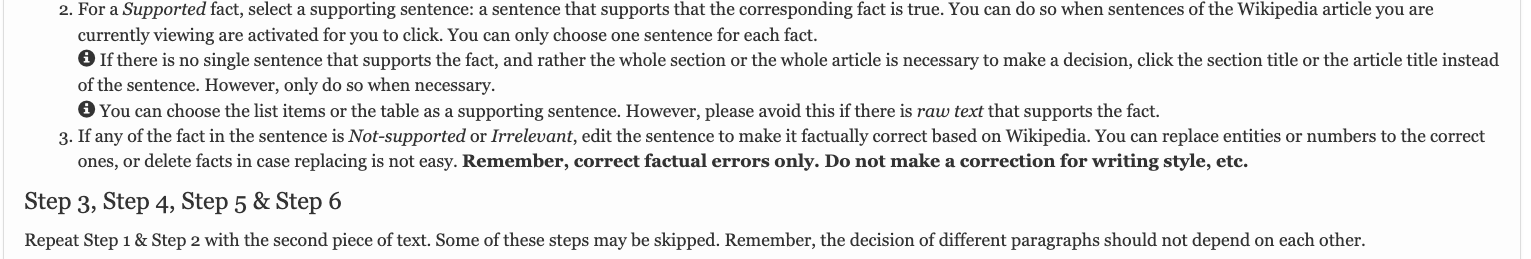}%
}
\resizebox{2\columnwidth}{!}{
    \includegraphics[trim={0 0 0 0},clip]{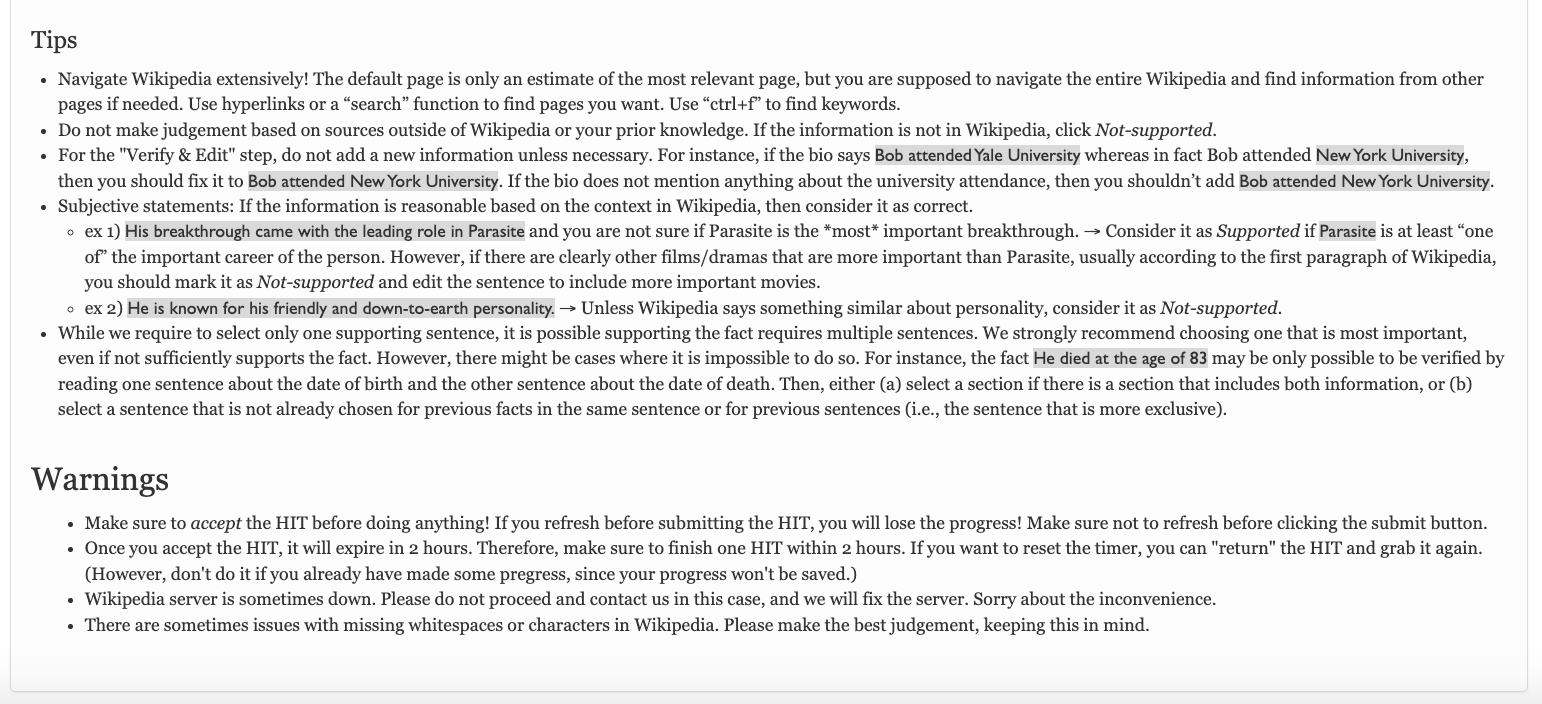}%
}
\vspace{-.3em}
\caption{Instructions for data annotation in Section~\ref{sec:method}. We also provided a demonstration video, and gave feedback 1-1 during the qualification task.
}\label{fig:instruction}
\end{figure*}

\begin{figure*}[t]
\centering \footnotesize
\resizebox{2\columnwidth}{!}{
    \includegraphics[trim={0 0 0 0},clip]{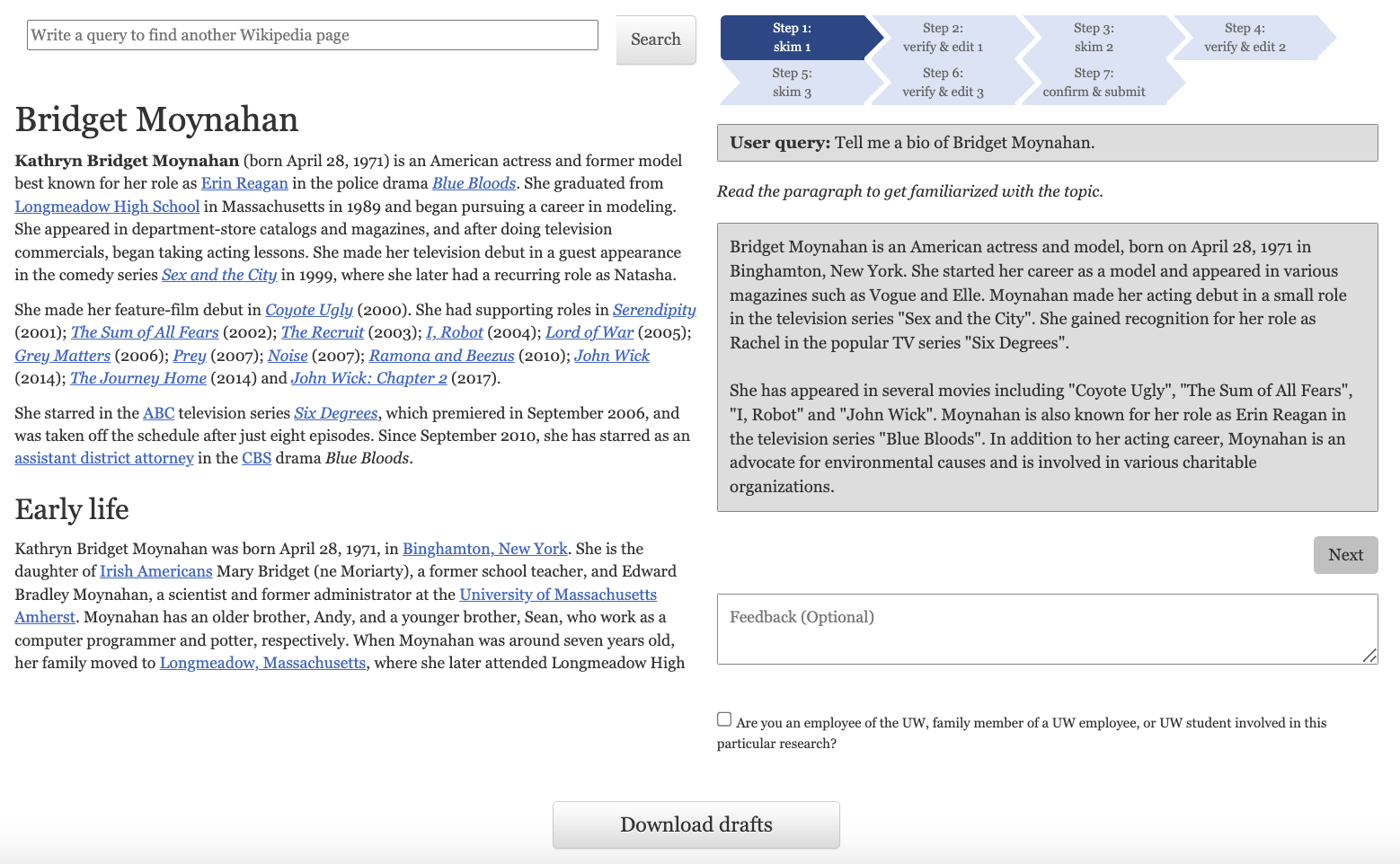}%
}
\resizebox{2\columnwidth}{!}{
    \includegraphics[trim={0 0 0 0},clip]{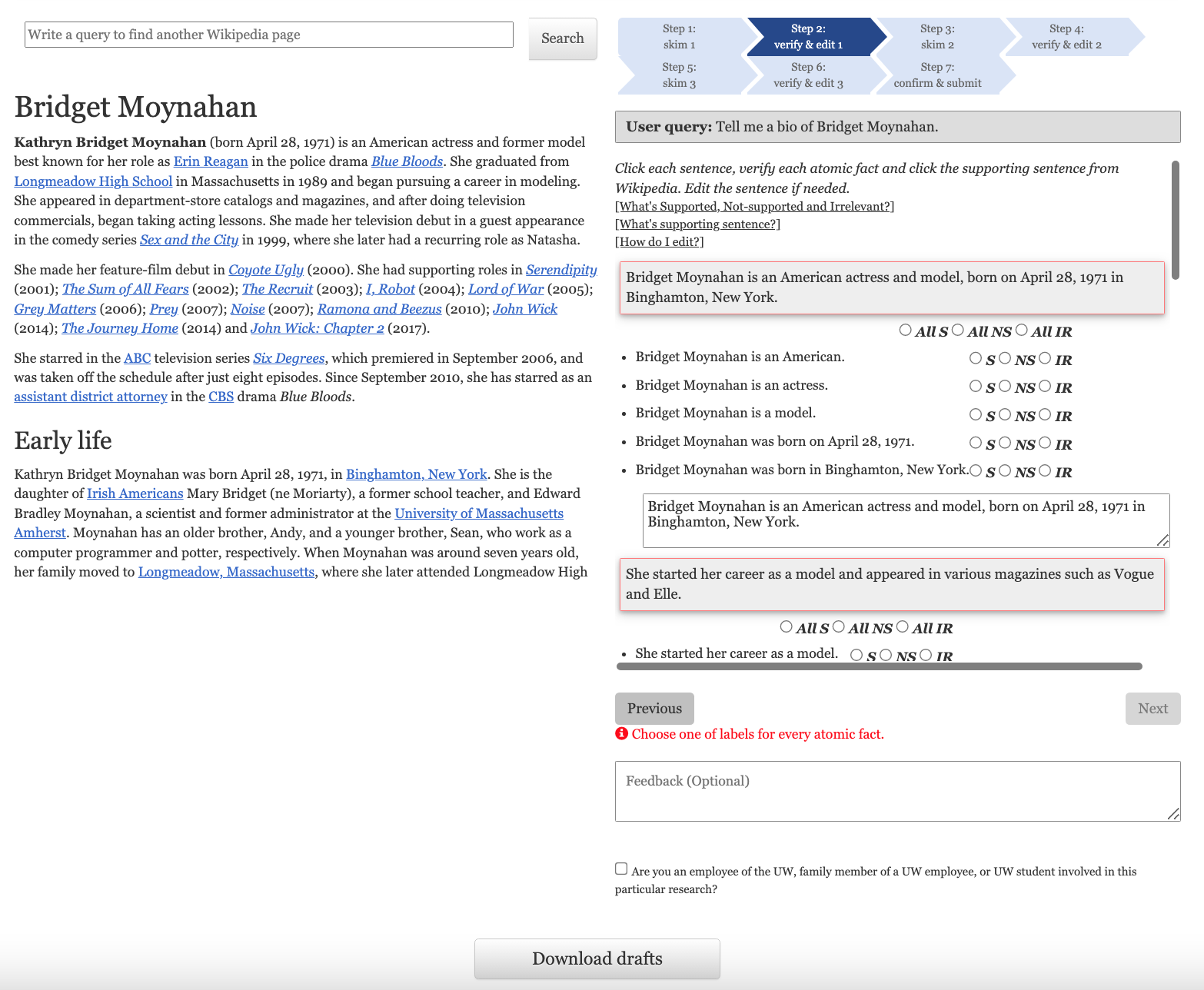}%
}
\vspace{-.3em}
\caption{An interface for data annotation in Section~\ref{sec:method}. Annotators were able to navigate Wikipedia on the left. They annotate three pieces of generations from three LMs for the same prompt in one HIT since it saves time. Since completing one HIT takes considerable amount of time (25min), we added a function that allows saving their work at any stage in the middle of the HIT.
}\label{fig:interface}
\end{figure*}

We recruit freelancers through Upwork and pay 15--25 USD per hour. We recruit fact-checking experts---freelancers who mentioned fact-checking as their expertise---for Step 3.
Every worker went through a qualification test of 2 hours and was tested to be highly qualified.
We design one HIT to consist of three generations, one from each \subjectLM, for one prompt, because we find it saves annotation time in total.
10\% of the HITs have two workers assigned to calculate the agreement rate; the rest have one worker assigned.
The agreement rates are 96\%, 90\% and 88\% for InstructGPT, ChatGPT and PerplexityAI, respectively. Appendix~\ref{subsec:annotation-qualitative} discusses disagreement cases in more detail.
The full instructions and the interface are provided in Figure~\ref{fig:instruction} and Figure~\ref{fig:interface}, respectively.


\begin{table}[t]
    \myfontsize \setlength{\tabcolsep}{0em}
    \centering
    \begin{tabular}{p{7.6cm}}
        \toprule
        \textbf{Prompt:} Tell me a bio of Ylona Garcia. \\
        \textbf{Sentence:} [Ylona Garcia] has since appeared in various TV shows such as ASAP (All-Star Sunday Afternoon Party), Wansapanataym Presents: Annika PINTAsera and Maalaala Mo Kaya. \\
        $\mathbf{\bullet}$ Ylona Garcia has appeared in various TV shows. \myblue{\sLabel} \\
        $\mathbf{\bullet}$ She has appeared in ASAP. \myblue{\sLabel}  \\
        $\mathbf{\bullet}$ ASAP stands for All-Star Sunday Afternoon Party. \myblue{\sLabel} \\
        $\mathbf{\bullet}$ ASAP is a TV show. \myblue{\sLabel}\\
        $\mathbf{\bullet}$ She has appeared in Wansapanataym Presents: Annika PINTAsera. \mypurple{\nsLabel}\\
        $\mathbf{\bullet}$ Wansapanataym Presents: Annika PINTAsera is a TV show.  \mypink{\irLabel} \\
        $\mathbf{\bullet}$ She has appeared in Maalaala Mo Kaya.  \mypurple{\nsLabel} \\
        $\mathbf{\bullet}$ Maalaala Mo Kaya is a TV show. \mypink{\irLabel} \\
        \midrule
        %
        \textbf{Prompt:} Tell me a bio of John Estes.\\
        \textbf{Sentence:} William Estes is an American actor known for his role on CBS police drama Blue Bloods as Jameson \"Jamie\" Reagan. \\
        $\mathbf{\bullet}$ William Estes is an American. \mypink{\irLabel} \\
        $\mathbf{\bullet}$ William Estes is an actor. \mypink{\irLabel}\\
        $\mathbf{\bullet}$ William Estes is known for his role on CBS police drama Blue Bloods. \mypink{\irLabel}\\
        $\mathbf{\bullet}$ William Estes' role on Blue Bloods is Jameson ``Jamie'' Reagan. \mypink{\irLabel} \\
        \bottomrule
    \end{tabular}
    \caption{
      Examples that contain \myblue{\sLabel}, \mypurple{\nsLabel} and \mypink{\irLabel}. Sentences in bullet points indicate atomic facts.
    }\label{tab:example-label}
\end{table}

\begin{table*}[t]
    \center  \myfontsize 
    \setlength{\tabcolsep}{0.5em}
    \begin{tabular}{p{3.3cm}p{0.4cm}p{11.4cm}}
        \toprule
            Category & \% & Example \\
        \midrule
            Different interpretations of the factual information
            & 21
            & \myblue{Gen} Gerhard Fischer is an inventor. \mypink{Wiki} Gerhard Fischer (inventor). ... was first patented by Dr. Gerhard Fischer  in 1931. A metal detector had been invented some forty years earlier (1881) by Alexander Graham Bell ... \\
            && \myblue{Gen} Chadwick Boseman was a producer. \mypurple{Comment} Chadwick Boseman is not known as a producer, but produced one music video. \\
        \cmidrule(lr){1-3}
            Inferred (not directly mentioned but highly likely)
            & 16
            & \myblue{Gen}  Leach has since become a member of the England Test team. \mypurple{Comment} Leach is a member of the England Test team, but since when is less clear. \\
        \cmidrule(lr){1-3}
            Depends on how strict in judging the correctness
            & 11
            & \myblue{Gen} He made his Test debut for England in March 2018. \mypink{Wiki} On 16 March 2018, he was called up to England's Test squad (...) He made his debut in the second Test in Christchurch. \\
            && \myblue{Gen} The building was the first LEED-certificated building in Edmonton. \mypink{Wiki} (..) became the first project in the City of Edmonton to achieve a LEED Gold status. \\
        \cmidrule(lr){1-3}
            Subjective  & 21 & 
            \myblue{Gen} Chadwick Boseman became an African American pioneer. \mypink{Wiki} Culture writer Steve Rose, in The Guardian, said that Boseman's career was revolutionary and he ``leaves behind a gamechanging legacy'' (...) Rose wrote: ``Chadwick Boseman began his career playing African American icons and pioneers; he ends it as one himself.'' \\
        \cmidrule(lr){1-3}
            Wikipedia not consistent & 5 &
            \myblue{Gen} [Tim Fischer] was an Ambassador to the Holy See from 2009 to 2012.
            \mypink{Wiki} ... was later Ambassador to the Holy See from 2009 to 2012. (...) Australian Ambassador to the Holy See 2008–2012 \mypurple{Comment} The plain text and the table of the {\em Tim Fischer} page as well as the {\em Australian Ambassador to the Holy See} page are inconsistent in his start year. \\
         \cmidrule(lr){1-3}
            Two different entities  & 5 & \mypurple{Comment} Carlos J. Alfonso vs. Carlos Alfonso \\
        \cmidrule(lr){1-3}
            Mistakes in annotation  & 21 & \myblue{Gen} Jack Leach is a left-handed batsman. \mypurple{Comment} mentioned in the {\em England cricket team} page, Table {\em Current Squad}.            \\
        \bottomrule
    \end{tabular}
    \caption{Categorization of disagreement cases.
     \myblue{Gen} indicates the generation from PerplexityAI, and \mypink{Wiki} indicates evidence text from Wikipedia. \mypurple{Comment} indicates our comments.
    }\label{tab:disagreement-analysis}
\end{table*}

\subsection{Examples in annotated data}\label{app:example-annotated-data}
Table~\ref{tab:example-label} provides examples of the human-annotated data, each atomic fact with an assigned label. \sLabel\ and \nsLabel\ respectively indicate Wikipedia supports the fact and does not support the fact (either contradicts or does not contain any evidence). \irLabel\ indicates the fact is irrelevant to the input prompt, which can further be divided into two cases: (1) the fact depends on other facts because it expands previous facts in a generation, and such other facts are \nsLabel, e.g., in the first example in Table~\ref{tab:example-label}, and (2) the entire sentence is irrelevant to the prompt, independent from other facts in a generation, e.g., the second example in Table~\ref{tab:example-label}.
The second case rarely happens with InstructGPT and ChatGPT, but happens considerably with PerplexityAI, i.e., 24.7\% of generations of PerplexityAI have $\geq$ sentences marked as irrelevant without dependencies to other facts, compared to 0.5\% and 1.3\% in InstructGPT and ChatGPT, respectively.
This is because PerplexityAI often directly copies search results even if they are largely irrelevant to the input prompt.
This is in agreement with a concurrent work from \citet{liu2023evaluating} that shows generative search engines like PerplexityAI copy incorrect search results and generate text that is irrelevant to the input query.

\subsection{Qualitative Analysis}
\label{subsec:annotation-qualitative}


\myskip{
\begin{table*}[t]
    \center \myfontsize
    \begin{tabular}{l cccc l
        }
        \toprule
             & \#Params & prompt types & retrv & need \subjectLM\ & closest prior work \\
        \midrule
            No-context LM   & 7B+ & \{\texttt{QA,FV}\} & \xmark & \xmark& \citet{kadavath2022language} \\
            Self-check LM   & 7B+ & \{\texttt{QA,FV}\} & \xmark & \cmark & \citet{manakul2023selfcheckgpt} \\
            \rtg            & 335M + 7B+ & \{\texttt{QA,FV}\} & \cmark & \xmark & \citet{gao2022attributed} \\
            NPM             & 354M & - & \cmark & \xmark & \citet{min2022nonparametric} \\
            \rtg\ + NPM     & 335M + 354M + 7B+ & \{\texttt{QA,FV}\} & \cmark & \xmark & - \\
        \bottomrule
    \end{tabular}
    \caption{Summary of five evaluators described in Section~\ref{subsec:models-validation}.
    7B+ is from LLAMA 7B and can be more if larger LMs are being used; 335M is from GTR large; 354M is from NPM.
    {\em retrv} indicates whether or not retrieval over an external corpus (Wikipedia) is being used.
    {\em need \subjectLM{}} indicates whether multiple samples from the \subjectLM{} are needed; methods with this requirement cannot be applied to the PerplexityAI-based data, because PerplexityAI is semi-deterministic (likely due to internal cache).
    }\label{tab:validation-methods-summary}
\end{table*}
}

\paragraph{Analysis of disagreement cases.}
We analyze the cases where two annotators assigned to a same generation disagree on a precision label for the same atomic fact. Categorization is provided in Table~\ref{tab:disagreement-analysis}. The 70\% is due to an inherent debatability on whether or not the fact is supported by a given source of knowledge, not satisfying Assumption 2 in Section~\ref{subsec:data-overview}.
This is because there can be multiple interpretations of a fact, it is debatable whether or not an information can be inferred from a piece of text, or the atomic fact is subjective. For instance:\vspace{-.3em}
\begin{itemize}[leftmargin=15pt]\itemsep -.2em
    \item \texttt{Gerhard Fischer is an inventor}: Gerhard Fischer is widely known as an inventor of a metal detector, and even the title of the Wikipedia article is \emph{``Gerhard Fischer (inventor)''.}
    However, it later turns out that he did not invent a metal detector; rather, he commercialized it.
    \item \texttt{Chadwick Boseman was a producer}: Chadwick Boseman is widely known as another profession (singer) and there is no text that mentions him as a producer. However, he produced one music video.
\end{itemize}
Nonetheless, since our agreement rate is fairly high (91\%), we think such cases are rare in our particular domain of people biographies. We include more discussion on other domains that such cases may be more frequent in the Limitation section.

\paragraph{Coverage of English Wikipedia.}
While factual prediction is inherently a function of a knowledge source given as part of the input, a potential concern is how representative using English Wikipedia as a knowledge source for evaluating people biographies with respect to its coverage. For instance, it is possible that, especially for rare entities, the coverage of information in Wikipedia is not high enough, and LMs may be penalized by generating information that is true even if not supported by Wikipedia (i.e., supported by other sources on the web).

To quantify the effect, we randomly sample 30 unsupported facts from ChatGPT on people whose categories are either `rare' or `very rare', and then validate them against the entire web. We found 10\% (3 out of 30 facts) are in fact supported, even though they are not supported in Wikipedia. An example is \emph{[Hibo] Wardere published her memoir titled ``Cut: One Woman's Fight Against FGM in Britain Today''} which is not mentioned in Wikipedia but is found from Google Books.

Nonetheless, we found that Wikipedia has a high coverage and mentions most of the important information that we were able to find from any other sources on the web. This is in agreement with prior work that treated Wikipedia as a general knowledge source under the same reason~\cite{chen-etal-2017-reading,petroni-etal-2021-kilt}.

\section{Details in Estimators}\label{app:exp-details}
\subsection{Implementation details}\label{subsubsec:setup-validation}

As an \evalLM, we use the best open LM and the best commercial LM at the time of conducting experiments: LLAMA 65B~\citep{touvron2023llama} and LLAMA 7B trained on Super Natural Instructions (Inst-LLAMA, \citealp{wang2022super}) as the former, and ChatGPT~\citep{chatgpt} as the latter. 
For computing nonparametric probabilities, we use a single-mask variant of NPM with BM25 as in the original paper~\citep{min2022nonparametric}, and use $0.3$ as a thresholding hyperparameter.

For passage retrieval, we use Generalizable T5-based Retrievers (GTR, a large variant), an unsupervised dense passage retrieval system~\citep{ni2021large}. We restrict retrieved passages to be from the topic entity's page, and use $k=5$.
We find our estimator is not sensitive to the choice of a retrieval system (ablations provided in Appendix~\ref{app:val-abl}). 
As a retrieval corpus, we use the English Wikipedia from 04/01/2023 which is around the time the data annotation was completed, and split each page into passages with up to 256 tokens.

\paragraph{Additional baselines.} We also compare with \textbf{Self-check LM}, a method from a concurrent work by \citet{manakul2023selfcheckgpt}. Self-check LM needs multiple samples generated from the \subjectLM. It validates the given atomic fact by prompting \evalLM\ conditioning on each generated sample,\footnote{\citet{manakul2023selfcheckgpt} uses BERTScore and a supervised question answering system instead of LM prompting, however, we find LM prompting to be significantly better.}  making judgment (\sLabel\ or not) from each, and aggregates the results through a majority vote.
This method assumes (1) the \subjectLM\ is available at the time of evaluation and (2) the outputs from the \subjectLM\ are nondeterministic, which makes it not applicable to PerplexityAI.

\subsection{Segment-level vs. system-level evaluation}\label{app:fone}

Besides how close the estimated \ours\ is to the ground truth \ours\ (\textbf{Error Rate}, as reported in Section~\ref{sec:method}), we also report \textbf{\Fone}. \Fone\ evaluates how well the model validates each individual atomic fact, assuming oracle atomic facts (atomic facts by human experts) are given, and evaluates how good the estimator is in identifying facts that are not \sLabel\ (\nsLabelShort). Formally, let $\mathcal{G}$ and $\mathcal{P}$ be sets of atomic facts in a set of generations that have \nsLabel\ as a ground truth label and as a predicted label, respectively. We define \Fone\ as follows.
\begin{align*}
\begin{gathered}
    \mathrm{P}=\frac{\mathcal{P}\cap\mathcal{G}}{\mathcal{P}},~~
    \mathrm{R}=\frac{\mathcal{P}\cap\mathcal{G}}{\mathcal{G}},~~\textbf{F1}_\textsc{micro}=\frac{2\cdot\mathrm{P}\cdot\mathrm{R}}{\mathrm{P}+\mathrm{R}} \\
\end{gathered}
\end{align*}
We call them \textsc{micro} because they consider individual decisions rather than aggregated estimation.

\begin{figure}[t]
\centering \footnotesize
\resizebox{0.9\columnwidth}{!}{
    \includegraphics[trim={36.5cm 7cm 9cm 27.5cm},clip]{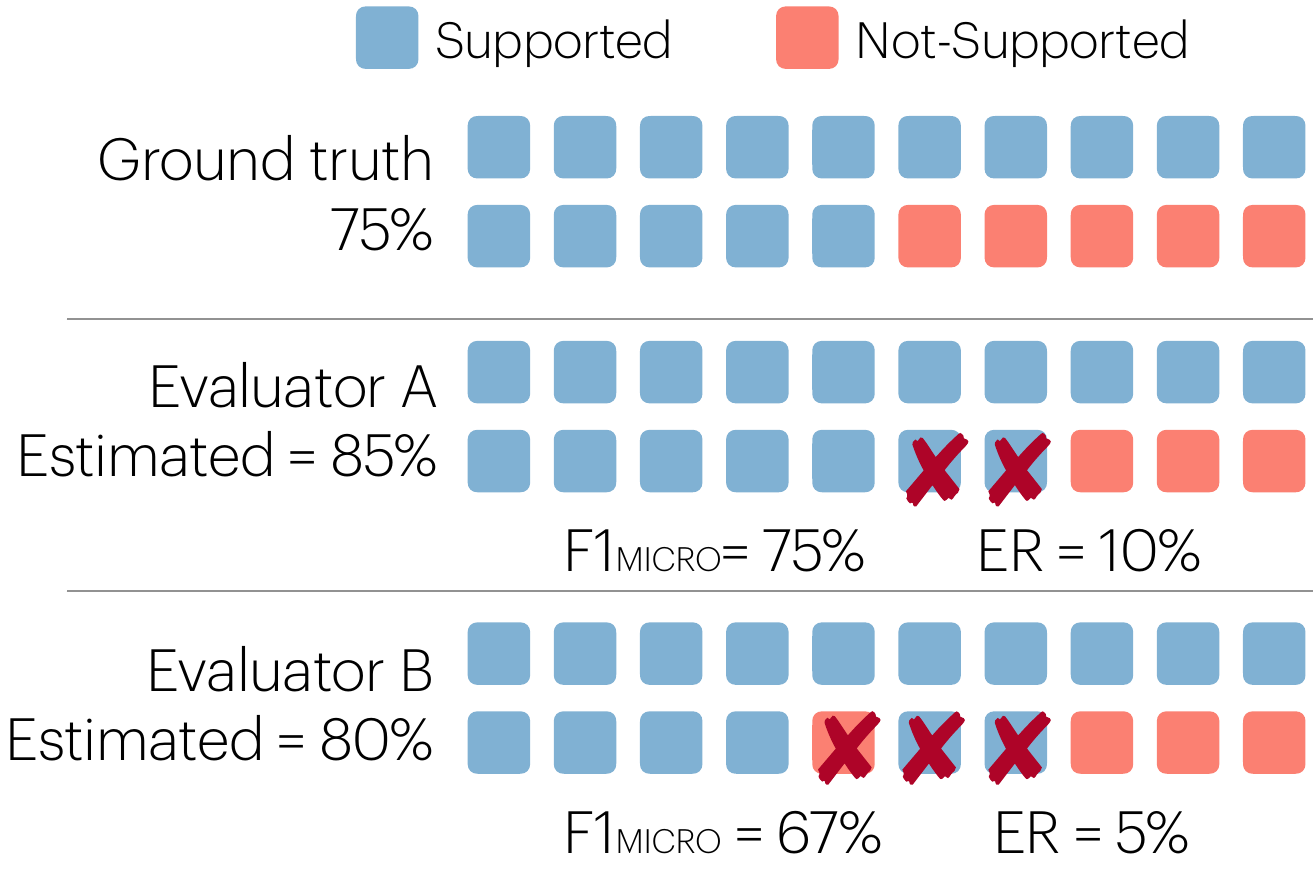}%
}\vspace{-.3em}
\caption{A case in which \Fone\ and Error Rate (ER) rank two evaluators differently. Evaluator A is better in \Fone, and Evaluator B is better in ER.}\label{fig:metrics}
\end{figure}

\paragraph{ER vs. \Fone.}
\Fone\ cares about the individual decision, while ER cares about the aggregated estimation.
An evaluator that has a high (better) \Fone\ but always overestimates or underestimates factual precision may have a higher (worse) ER, e.g., Evaluator A in Figure~\ref{fig:metrics}.
Conversely, an evaluator that has a lower (worse) \Fone\ but is not biased toward overestimation nor underestimation may have a lower (better) ER, e.g., Evaluator B in Figure~\ref{fig:metrics}.
Prior work in model-based evaluation mainly reports aggregated scores since the goal is a comparison between different systems being evaluated~\citep{zhang2019bertscore,rashkin2021measuring,gao2022attributed} while we report both to see the relationship between two types of metrics. 
\Fone\ and ER are also closely related to  \emph{segment}-level and \emph{system}-level correlations to human judgments respectively, which have been extensively used in developing evaluation metrics in machine translation~\citep{ma-etal-2019-results,thompson-post-2020-automatic} and summarization~\citep{bhandari-etal-2020-evaluating,deutsch2021towards}.

\begin{table}[t]
    \center  \footnotesize 
    \setlength{\tabcolsep}{0.5em}
    \begin{tabular}{l @{\hspace{-0.7em}}
            R{0.7cm}
            @{\hspace{0.3em}} R{1.2cm}
            @{\hspace{0.2em}} R{1.2cm}
            @{\hspace{0.2em}} R{1.2cm}
        }
        \toprule
            \multirow{2}{*}{Evaluator} & \multirow{2}{*}{retrv} & \multicolumn{3}{c}{\subjectLM} \\
            \cmidrule(lr){3-5}
            &&{InstGPT} & {ChatGPT} & {PPLAI} \\
        \midrule        
            Always \sLabel & -      &  0.0 & 0.0  & 0.0  \\
            Always \nsLabel & -     & 71.4  & 58.3 & 30.9  \\
            Random & -              & 52.2  & 45.0  & 25.7  \\
        \midrule
            No-context LM &\xmark   & 61.2  & 52.2 & 31.4 \\
            Self-check LM &\xmark   & 66.0 & 48.4 & - \\
        \midrule
            \rtg        &\cmark     & 78.7  & 61.9 & 51.1\\
            NP         &\cmark     & 70.0  & 56.6  & 51.4 \\
            \rtg\ + NP  &\cmark    & \textbf{83.2}  & \textbf{70.5} & \textbf{53.3} \\
        \bottomrule
    \end{tabular}
    \caption{
        Results in \textbf{\Fone} using Inst-LLAMA 7B 
        as an \evalLM. 
        `{\em retrv}' indicates whether or not retrieval is used.
        Self-check is not applicable to PerplexityAI whose outputs are semi-deterministic. \textbf{Bold} indicates the best performance.
    }\label{tab:validation-results}
\end{table}

\paragraph{Results.} Results on \Fone\ are reported in Table~\ref{tab:validation-results}.
Self-check LM outperforms no-context LM by 4--11\%, which confirms findings from \citet{manakul2023selfcheckgpt}.
However, both significantly underperform methods that use retrieval. This is in contrast to \citet{manakul2023selfcheckgpt} that reports that Self-check without retrieval achieves performance that is close to that with retrieval, likely because the data in \citet{manakul2023selfcheckgpt} contains more frequent entities.
The fact that retrieval significantly helps is consistent with findings in Section~\ref{subsec:models-validation-result} with an ER as a metric.

Adding NP improves \rtg\ by 2--9\%, again consistent with findings in Section~\ref{subsec:models-validation-result}. This is likely because \rtg\ often makes incorrect predictions when there is a strong bias from an LM or there are distracting passages, and considering nonparametric probabilities makes the model more robust to these factors. For instance, given an unsupported fact \texttt{Samuel Oboh is Nigerian}, No-context LM, Self-check LM and \rtg\ predict \sLabel\ due to a strong name-nationality bias.
NPM correctly predicts \nsLabel\ based on a passage \texttt{Samuel Oboh ... is a Canadian architect, manager, ...}.
It is also worth noting that this is different from findings in Section~\ref{subsec:models-validation-result} that ChatGPT is not necessarily better than LLAMA+NP  based on ER.

\paragraph{Using a stronger \evalLM\ significantly improves \Fone.}
Table~\ref{tab:validation-results-ablation} reports a comparison across different choices of an \evalLM{}. Within the same method, Inst-LLAMA 7B outperforms LLAMA 65B, and ChatGPT outperforms both.
Using retrieval is critical across all models, e.g., the best no-context model based on ChatGPT is underperformed by all models with retrieval.
Using NP helps LLAMA-based models but not ChatGPT, likely because ChatGPT is less affected by incorrect prior from the LM or distracting passages.

It is worth noting that these results are somewhat different from findings in Section~\ref{subsec:models-validation-result} that ChatGPT is not necessarily better than LLAMA+NP. This is becauase, although ChatGPT is better in validating each individual atomic fact, most errors from ChatGPT are incorrectly assigning \sLabel\ to \nsLabel\ facts, resulting in an overestimation of \ours.
In contrast, LLAMA+NP is not biased toward overestimation or underestimation of the factual precision, resulting in an aggregated factual precision to be closer to the ground truth.
This is similar to the trade-off between system-level and segment-level correlations in summarization evaluation~\citep{bhandari-etal-2020-evaluating,deutsch2021towards}.

\begin{table}[t]
    \center \footnotesize 
    \setlength{\tabcolsep}{0.5em}
    \begin{tabular}{l @{\hspace{-0.7em}}
            R{0.7cm}
            @{\hspace{0.3em}} R{1.2cm}
            @{\hspace{0.2em}} R{1.2cm}
            @{\hspace{0.2em}} R{1.2cm}
        }
        \toprule
            \multirow{2}{*}{Evaluator} & \multirow{2}{*}{retrv} & \multicolumn{3}{c}{\subjectLM} \\
            \cmidrule(lr){3-5}
            &&{InstGPT} & {ChatGPT} & {PPLAI} \\
        \midrule 
            \multicolumn{5}{l}{\textbf{\em LLAMA 65B}} \\
            No-context LM & \xmark  & 22.2 & 20.0 & 18.6 \\
            \rtg\ & \cmark          & 54.6 & 42.1 & 36.1 \\
            \rtg\ + NP & \cmark    & 80.1 & 67.1 & \textbf{55.1} \\
        \midrule
            \multicolumn{5}{l}{\textbf{\em Inst-LLAMA 7B}} \\
            No-context LM &\xmark   & 61.2 & 52.2 & 31.4 \\
            \rtg & \cmark           & 78.7 & 61.9 & 51.1\\
            \rtg\ + NP & \cmark    &\textbf{83.2} &\textbf{70.5}& 53.3 \\
        \midrule
            \multicolumn{5}{l}{\textbf{\em ChatGPT}} \\
            No-context LM & \xmark  & 40.0 & 25.4 & 25.4\\
            \rtg & \cmark           & \best{87.5} & \best{80.2} & \best{65.8} \\
            \rtg\ + NP & \cmark    & 86.6 & 77.8 & 60.8 \\
        \bottomrule
    \end{tabular}
    \caption{Ablation in \textbf{\Fone} on the choices of \evalLM. `{\em retrv}' indicates whether or not retrieval is used.
    \textbf{Bold} and \best{Red bold} indicate the best F1 within open-access LMs and commercial LMs, respectively. 
    }\label{tab:validation-results-ablation}
\end{table}

\subsection{Ablations}\label{app:val-abl}

\paragraph{\texttt{QA} Prompting vs. \texttt{TF} Prompting}
As described in Section~\ref{subsec:models-validation}, we use \texttt{True or False} as part of the prompt, so-called \texttt{TF} Prompting. An alternative is \texttt{QA} Prompting, which generates a question and the expected answer, obtains the answer for the generated question independent from the expected answer, and compares the expected answer and the predicted answer. This approach has been widely studied in the summarization literature and recent work in factual precision~\citep{kryscinski2019evaluating,wang2020asking,gao2022attributed,manakul2023selfcheckgpt}.
Table~\ref{tab:abl-prompting} provides a comparison between two types of prompting.
The \texttt{TF} approach significantly outperforms the \texttt{QA} approach, consistently over all methods. Our further analysis finds that this is due to generated questions often being overly vague or ambiguous. For instance, given a supported fact \texttt{Samuel Oboh is an architect}, the LM generates \texttt{What is Samuel Oboh's job?} as a question and \texttt{Architect} as an expected answer, and the obtained answer is \texttt{Vice President}. Although both \texttt{Architect} and \texttt{Vice President} are correct, they are not the same, thus the model incorrectly predicts \nsLabel.
Such cases make the model overpredict \nsLabel, leading to many incorrect predictions.

\begin{table}[t]
    \center \footnotesize
    \begin{tabular}{l
            @{\hspace{0.2em}} R{1.3cm}
            @{\hspace{0.2em}} R{1.3cm}
            @{\hspace{0.2em}} R{1.3cm}
        }
        \toprule
            \multirow{2}{*}{Evaluator} & \multicolumn{3}{c}{\subjectLM} \\
            \cmidrule(lr){2-4}
            & InstGPT & ChatGPT & PPLAI \\
        \midrule        
            Always \sLabel & \alwaysS \\
            Always \nsLabel & \alwaysNS \\
            Random & \alwaysRandom \\
        \midrule
            \multicolumn{4}{l}{\em \textbf{\texttt{QA} Prompting}} \\
            No-context LM  & \noContextQA \\
            Self-check LM & \selfcheckQA \\
            \rtg          & 65.3 & 58.2 & 47.3 \\
        \midrule
            \multicolumn{4}{l}{\em \textbf{\texttt{TF} Prompting}} \\
            No-context LM & \noContextFV \\
            Self-check LM & \selfcheckFV \\
            \rtg        & \textbf{78.9} & \textbf{71.4} & \textbf{69.2} \\
        \bottomrule
    \end{tabular}
    \caption{
        Results on \Fone, comparing between the \texttt{QA} prompting and \texttt{TF} Prompting.
        We use Inst-LLAMA 7B as an \evalLM.
        Self-check is not applicable to PerplexityAI since PerplexityAI outputs are semi-deterministic.
        \textbf{Bold} indicates the best \Fone.
    }\label{tab:abl-prompting}
\end{table}

\begin{table}[t]
    \center \footnotesize
    \begin{tabular}{l
            @{\hspace{0.2em}} R{1.3cm}
            @{\hspace{0.2em}} R{1.3cm}
            @{\hspace{0.2em}} R{1.3cm}
        }
        \toprule
            \multirow{2}{*}{Retrieval} & \multicolumn{3}{c}{\subjectLM} \\
            \cmidrule(lr){2-4}
            & InstGPT & ChatGPT & PPLAI \\
        \midrule 
            BM25 & 78.5 & 70.8 & 69.1\\
            GTR Large & 78.9 & \textbf{71.4} & \textbf{69.2} \\
            GTR xLarge & \textbf{79.2} & 71.3 & 69.0 \\
        \bottomrule
    \end{tabular}
    \caption{
        Results on \Fone, comparing different retrieval systems: BM25, GTR Large and GTR xLarge, all with \rtg\ based on Inst-LLAMA 7B.
        \textbf{Bold} indicates the best \Fone.
    }\label{tab:abl-retrieval}
\end{table}

\begin{table}[t]
    \center \footnotesize
    \begin{tabular}{lr
        }
        \toprule 
            Category & \% \\
        \midrule
            No direct evidence from retrieved passages & 70 \\
            Distracted by other passages & 17 \\
            Atomic fact is context-dependent & 7 \\
            Wrong prediction even with the right passage & 3 \\
            Annotation error & 3 \\
        \bottomrule
    \end{tabular}
    \caption{
        Categorization of 30 samples incorrectly predicted by \rtg\ based on ChatGPT.
    }\label{tab:chatgpt-errors}
\end{table}

\vspace{-.3em}
\paragraph{Impact of the choice of retrieval.}
Table~\ref{tab:abl-retrieval} compares \rtg\ methods based on a few passage retrieval systems, including BM25~\citep{Lin_etal_SIGIR2021_Pyserini}, GTR Large and GTR xLarge. 
Results indicate that all retrieval systems are equally good and \rtg\ is not sensitive to the choice of the retrieval system.

\begin{figure*}[t]
\centering \footnotesize
\resizebox{2.1\columnwidth}{!}{
    \includegraphics[trim={0.3cm 0 0 0},clip]{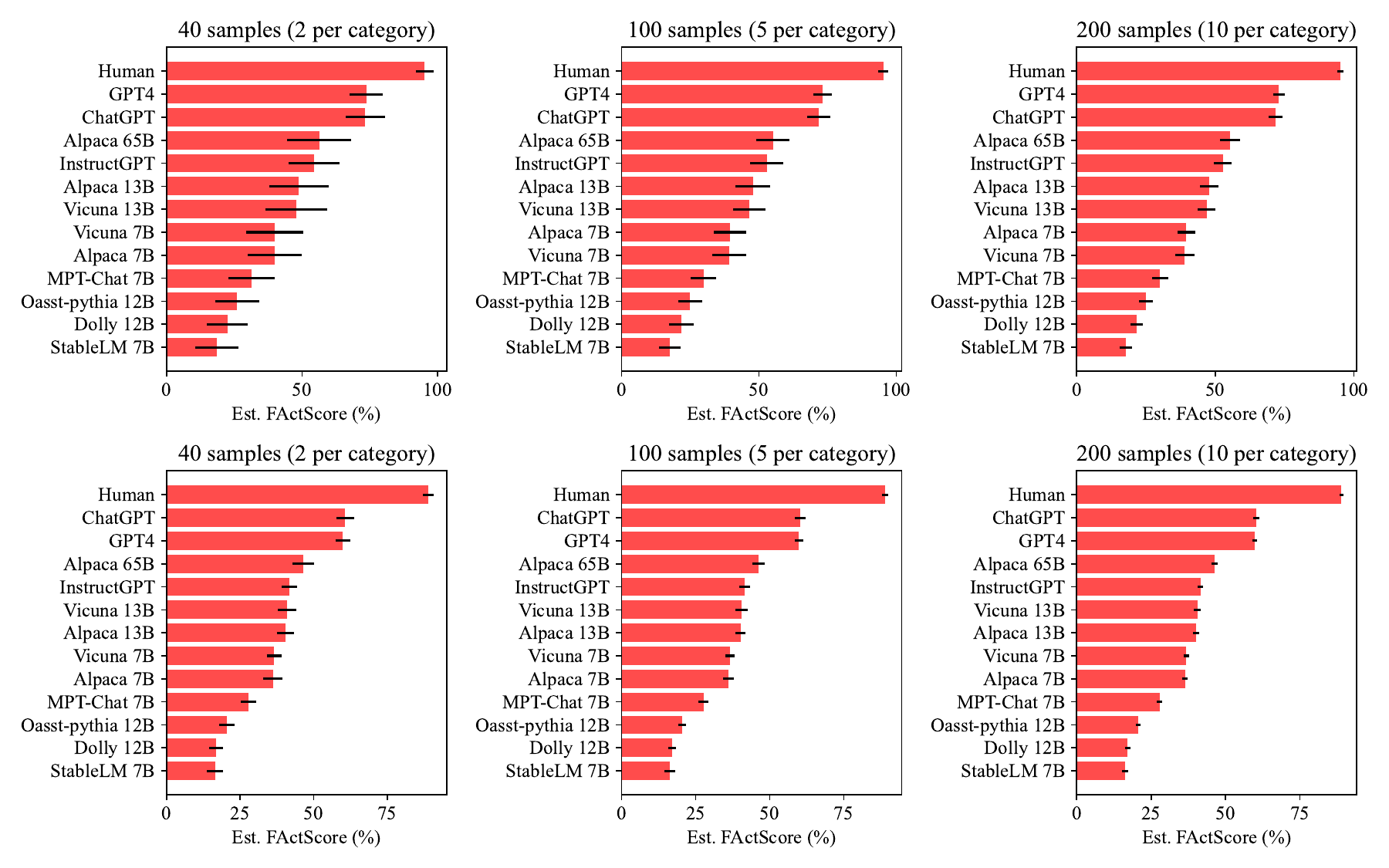}%
}\vspace{-.3em}
\caption{
    Impact of different subsets of random samples in prompts.
    The \ourScores\ to 13 subjects (human and 12 LMs) are rated by the two best variants of our estimator: ChatGPT (\textbf{Top}) and LLAMA+NP (\textbf{Bottom}), both with retrieval. The variance is overall low, and is lower as the sample size gets larger and with LLAMA+NP (bottom) than with ChatGPT (top).
    }\label{fig:new-lms-ranking-error-bar}
\end{figure*}

\vspace{-.3em}
\paragraph{Qualitative analysis.}
Table~\ref{tab:chatgpt-errors} categories errors made by \rtg\ based on ChatGPT, the evaluator with the best \Fone. 70\% of the errors are due to retrieved passages not providing direct evidence (either support or contradiction).
These are difficult even for state-of-the-art retrieval systems and language models because validating facts often requires reading the entire page rather than a single passage, e.g., an actor not appearing in a particular film.
17\% of errors are made because ChatGPT is being distracted by other passages, although it assigns a correct label if only a particular, correct passage is given.

\subsection{More details in evaluation of new LMs (Section~\ref{sec:eval-new-lms})}\label{app:new-lms-additional}

\paragraph{Variance in estimation.}
Figure~\ref{fig:new-lms-ranking-error-bar} reports \ourScores\ estimated by two variants of our estimator as in Figure~\ref{fig:new-lms-ranking} but with 100 random subsets of the data. Specifically, we chose $N$ samples (out of $500$) uniformly at random across 20 categories (defined in Appendix~\ref{app:data-category}) $M$ times and report the average and the standard deviation. We use $N=\{40, 100, 200\}$ and $M=100$. Results indicate that the variance is overall low, preserving ranking between 13 subjects in most cases. As expected, the variance is lower as the sample size gets larger. Finally, the estimator based on ER based on LLAMA+NP (bottom) has an overall lower variance than the estimator based on ChatGPT (top).

\subsection{Feasibility in applying \ours\ to other domains}\label{app:nlp-domain}
As mentioned in the Limitation section, our paper mainly evaluates on people biographies using Wikipedia. Evaluating the generalizability of \ours\ to other types of prompts and other domains is an avenue for future work.

As a proof of conept, we conduct small-scale studies in the NLP domain. We first manually write 10 prompts asking about NLP papers: \emph{Tell me a summary of}~\texttt{<paper-title>}, and then obtain responses from ChatGPT. Next, we run \ours\ against an ACL anthology as a knowledge source. Finally, we compute an error rate (ER)---a difference between humans’ validation (labeled by authors) and the model’s validation---as we do in Section~\ref{sec:method}. The ER is 7.41 (\ours\ from humans being 66.20, and \ours\ from the model being 73.61), which is comparable to ER values in people bios shown in Table~\ref{tab:validation-error-rate}. 

This suggests that \ours\ can generalize beyond people biographies. However, since this is a very small-scale experiment, we strongly encourage future research to explore the generalizability of \ours\ to more domains at scale.

\section{Editing Experiments}\label{sec:editing}
Our experiments in Section~\ref{sec:method} focuses on automatically identifying factual precision errors in long-form generations by language models. Can these labels be used to actually correct errors in the long-form generations? In this section, we perform a preliminary exploration of methods to edit long-form LM generations to reflect factually correct information. We assume we have access to the human-annotated set of \ours\ labels, and measure how good models are at editing incorrect sentences. In other words, we evaluate our editor models independent of the errors arising from the estimator. 

\subsection{Methods}\label{subsubsec:editing-methods}

We adopt a similar set of methods as Section~\ref{subsec:models-validation} for our editing models. All methods below use four exemplar examples for in-context learning which were sampled from our dataset and removed for subsequent analysis. For all methods, we use OpenAI's ChatGPT~\citep{chatgpt} as the base language model due to its generative capabilities.
\vspace{.3em}

\noindent \textbf{No-context LM}. We feed language models the prompt \texttt{Input: <sentence> Edit:} and ask it to edit the text, without any retrieved context. 

\vspace{.3em}

\noindent \textbf{\rtgShort}. To assist an editor model, we use a passage retrieval system to find supporting evidence from an external knowledge source (Wikipedia in our case). Our retrieval pipeline is identical to Appendix~\ref{subsubsec:setup-validation}, but uses 3 retrieved passages instead of 5 due to context length restrictions.

\vspace{.3em}

\noindent \textbf{+ Atomic Facts}. Additionally, we explore whether adding atomic facts and their labels assist a model with fine-grained editing. Specifically, after the input sentence we add information to the prompt of the form \texttt{Fact 1 (True/False): <atomic fact 1> Fact 2 (True/False): <atomic fact 2> ..}. This data is also provided in the exemplars.

\vspace{.3em}

\noindent \textbf{Non-edit baselines}. Finally, we add some trivial baselines to lower-bound our editing metrics. Specifically, we measure the performance of input copying (no edits), as well as an editor with random token dropping / replacement on a random 25\% subset of tokens.


\begin{table*}[t]
    \center \footnotesize
    \begin{tabular}{
            lrrrrrrrrr
        }
        \toprule
         & \multicolumn{3}{c}{InstructGPT} & \multicolumn{3}{c}{ChatGPT} & \multicolumn{3}{c}{PerplexityAI} \\
         \cmidrule(lr){2-4} \cmidrule(lr){5-7} \cmidrule(lr){8-10}
        Editor & ErrLoc & ErrCorr & SimAl & ErrLoc & ErrCorr & SimAl & ErrLoc & ErrCorr & SimAl  \\ 
        \midrule
        Input copying & 37.1 & 0.0 & 0.0 & 38.8 & 0.0 & 0.0 & 45.6 & 0.0 & 0.0 \\
        25\% random noise & 44.1 & 0.1 & 0.5 & 45.5 & 0.1 & 0.4 & 45.2 & 0.0 & 0.3  \\
        \midrule
        \multicolumn{5}{l}{\textbf{\em ChatGPT}} \\
        No-context & 49.0 & 8.5  &  6.2 & 45.3 & 6.8 & 4.0 & 48.3 & 6.2 & 4.1 \\
        No-context + atomic facts & 58.7 & 12.7 & 10.5 & 53.4 & 10.0  & 6.6 & 56.0 & 9.6 & 6.1 \\
        \rtgShort\ & 52.6 & 21.8 & 15.7 & 43.9 & 16.8 & 9.5 & 46.3 & 13.5 & 6.8 \\
        \rtgShort\ + atomic facts & \textbf{65.4} & \textbf{30.4} & \textbf{25.5} & \textbf{63.5} & \textbf{28.3}  & \textbf{19.3} & \textbf{62.4} & \textbf{23.6} & \textbf{15.9} \\
        \bottomrule
    \end{tabular}
    \caption{Results after automatic editing with ChatGPT assuming ground truth verification labels. All editors perform better than trivial lowerbound baselines, and using retrieval and atomic fact labels boosts editing performance. Details of automatic metrics (ErrLoc, ErrCorr, SimAl) are defined in Section~\ref{subsubsec:editing-eval}. 
    }\label{tab:editing-results}
\end{table*}

\subsection{Evaluation}\label{subsubsec:editing-eval}

In our data collection process (Section~\ref{subsec:data}), along with our verification data we also collected gold-standard human written edits. Let $X = x_1,...x_{N_X}$ be the input sentence and $G = g_1,...g_{N_G}$ be the gold edited sentence.
We evaluate the quality of the model-generated edit ($E = e_1,...,e_{N_E}$) using three automatic metrics,

\vspace{0.3em}

\noindent (1) \textbf{Error Localization} (ErrLoc): Our first metric measures how well the editor identifies errors within the input sentence. Specifically, we first create a ``token preservation string'', marking token $x_i$ in the input sentence $X$ as "Preserved" or "Not Preserved". We then compute the macro-averaged F1 score between the token preservation strings derived from the gold edit and the model-generated edit. We remove stopwords, punctuation and lowercase all words before performing this calculation. To equally weigh every sentence, F1 scores are independently computed for each sentence before a final averaging.

\vspace{0.3em}

\noindent (2) \textbf{Edit Correctness} (EditCorr): Our second metric assesses the quality of the additional tokens added by the model-generated edit. Specifically, we check the token-level F1 score~\citep{rajpurkar2016squad} comparing the new tokens added by the gold edit $G$ and the new tokens added by the model-generated edit $E$. More concretely,
\begin{align*}
N_\text{common} &= \sum_{e_i \in E, e_i \notin X} e_i \in G \\
\text{precision} &= N_\text{common} ~/~ || \{e_i \in E, e_i \notin X\} || \\
\text{recall} &= N_\text{common} ~/~ || \{g_i \in G, g_i \notin X\} || \\
\text{EditCorr (F1)} &= \text{HM} (\text{precision}, \text{recall})
\end{align*}

where $||\cdot||$ is the set cardinality and HM denotes a harmonic mean. For this metric, we discard data points where the gold edit did not add new tokens. Similar to ErrLoc, we also remove stopwords, remove punctuation and lowercase strings before calculating EditCorr scores.

\vspace{.3em}

\noindent (3) \textbf{SIM alignment} (SimAl): Finally, due to the large output space of possible edits, we also adopt a metric which rewards paraphrases of the gold edits. We use semantic similarity embeddings from~\citet{wieting-etal-2022-paraphrastic} which map paraphrases to a similar part of a vector space. We check the similarity between the model edit $E$ and the gold edit $G$, normalizing it by the similarity between $G$ and the original input $X$.\footnote{We avoid taking the vector differences between the original / edited text since edit vectors~\citep{guu2018generating} were not explicitly modeled in~\citet{wieting-etal-2022-paraphrastic}.} Specifically,
\begin{align*}
    \text{Sim} &= \max\left(0, \frac{s(G, E) - s(G, X)}{1 - s(G, X)} \right)
\end{align*}

where $s(A,B)$ is the semantic similarity score (normalized to $[0, 1]$) from the model in~\citet{wieting-etal-2022-paraphrastic}. Intuitively, this metric measures how much closer $G$ and $E$ are compared to $G$ and $X$.

\subsection{Results}\label{subsubsec:editing-results}

We present our editing results in Table~\ref{tab:editing-results}. Overall, we find that:

\vspace{.3em}

\noindent \textbf{All editing models perform better than trivial lower bounds.} Overall, we find that all editor models outperform lower-bound baselines like random noise. This even happens in the no-context LM setting, where ChatGPT is editing its own output (or search engine augmented Perplexity AI's outputs), but can still perform non-trivial corrections (6.8 ErrCorr for ChatGPT correcting its own outputs vs 0.1 for a random noise editor baseline).


\vspace{.3em}

\noindent \textbf{Retrieval significantly helps with editing performance.} Across all base language models and metrics, augmenting the editor with retrieved paragraphs boosts performance (6.8 $\rightarrow$ 16.8 ErrCorr, 4.0 $\rightarrow$ 9.5 SimAl for ChatGPT correcting its own outputs). We hypothesize that the internal parametric knowledge in ChatGPT has insufficient information about the topic (as we also observed in Section~\ref{subsec:annotation-results}) to perform fine-grained editing, and using external knowledge from Wikipedia greatly simplifies error localization and correction. This also corroborates with our findings in Section~\ref{subsec:models-validation-result}.

\vspace{.3em}

\noindent \textbf{Atomic fact labels improve error localization and improve editing performance.} Across all base language models (with or without retrieval) we observe that providing fine-grained atomic fact labels improves editing performance (16.8 $\rightarrow$ 28.3 ErrCorr, 9.5 $\rightarrow$ 19.3 SimAl for ChatGPT correcting its own outputs). Fine-grained fact correctness labels help the editor easily identify problematic tokens, as seen by the consistent improvements in ErrLoc scores (43.9 $\rightarrow$ 63.5 for ChatGPT correcting itself). We hypothesize atomic facts help guide the editor with its editing process (for instance, perform a more targeted search in the retrieved paragraphs), resulting in ErrCorr improvements. We also find that atomic fact labels reduces the frequency of editor copying the input verbatim or saying \emph{The input has no errors} from 37.3\% to 3.9\%.

\vspace{.3em}

\noindent \textbf{PerplexityAI outputs are the hardest to edit}. Overall, we find the highest editing success for InstructGPT, followed by ChatGPT and the least success for Perplexity AI. We hypothesize this is because PerplexityAI already uses a search engine, so errors are much more subtle as extensively discussed in Appendix~\ref{subsec:annotation-qualitative}.



\begin{table*}[t]
    \scriptsize \centering
    \begin{tabular}{|p{2\columnwidth}|}
        \toprule
Please breakdown the following sentence into independent facts: He made his acting debut in the film The Moon is the Sun's Dream (1992), and continued to appear in small and supporting roles throughout the 1990s. \\
- He made his acting debut in the film. \\
- He made his acting debut in The Moon is the Sun's Dream. \\
- The Moon is the Sun's Dream is a film. \\
- The Moon is the Sun's Dream was released in 1992.\\
- After his acting debut, he appeared in small and supporting roles.\\
- After his acting debut, he appeared in small and supporting roles throughout the 1990s.\\
\\
Please breakdown the following sentence into independent facts: He is also a successful producer and engineer, having worked with a wide variety of artists, including Willie Nelson, Tim McGraw, and Taylor Swift.\\
- He is successful.\\
- He is a producer.\\
- He is a engineer.\\
- He has worked with a wide variety of artists.\\
- Willie Nelson is an artist.\\
- He has worked with Willie Nelson.\\
- Tim McGraw is an artist.\\
- He has worked with Tim McGraw.\\
- Taylor Swift is an artist.\\
- He has worked with Taylor Swift.\\
\\
Please breakdown the following sentence into independent facts: In 1963, Collins became one of the third group of astronauts selected by NASA and he served as the back-up Command Module Pilot for the Gemini 7 mission.\\
- Collins became an astronaut.\\
- Collins became one of the third group of astronauts.\\
- Collins became one of the third group of astronauts selected.\\
- Collins became one of the third group of astronauts selected by NASA.\\
- Collins became one of the third group of astronauts selected by NASA in 1963.\\
- He served as the Command Module Pilot.\\
- He served as the back-up Command Module Pilot.\\
- He served as the Command Module Pilot for the Gemini 7 mission.\\
\\
Please breakdown the following sentence into independent facts: In addition to his acting roles, Bateman has written and directed two short films and is currently in development on his feature debut.\\
- Bateman has acting roles.\\
- Bateman has written two short films.\\
- Bateman has directed two short films.\\
- Bateman has written and directed two short films.\\
- Bateman is currently in development on his feature debut.\\
\\
Please breakdown the following sentence into independent facts: Michael Collins (born October 31, 1930) is a retired American astronaut and test pilot who was the Command Module Pilot for the Apollo 11 mission in 1969.\\
- Michael Collins was born on October 31, 1930.\\
- Michael Collins is retired.\\
- Michael Collins is an American.\\
- Michael Collins was an astronaut.\\
- Michael Collins was a test pilot.\\
- Michael Collins was the Command Module Pilot.\\
- Michael Collins was the Command Module Pilot for the Apollo 11 mission.\\
- Michael Collins was the Command Module Pilot for the Apollo 11 mission in 1969.\\
\\
Please breakdown the following sentence into independent facts: He was an American composer, conductor, and musical director.\\
- He was an American.\\
- He was a composer.\\
- He was a conductor.\\
- He was a musical director.\\
\\
Please breakdown the following sentence into independent facts: She currently stars in the romantic comedy series, Love and Destiny, which premiered in 2019.\\
- She currently stars in Love and Destiny.\\
- Love and Destiny is a romantic comedy series.\\
- Love and Destiny premiered in 2019. \\
\\
Please breakdown the following sentence into independent facts: During his professional career, McCoy played for the Broncos, the San Diego Chargers, the Minnesota Vikings, and the Jacksonville Jaguars.\\
- McCoy played for the Broncos.\\
- McCoy played for the Broncos during his professional career.\\
- McCoy played for the San Diego Chargers.\\
- McCoy played for the San Diego Chargers during his professional career.\\
- McCoy played for the Minnesota Vikings.\\
- McCoy played for the Minnesota Vikings during his professional career.\\
- McCoy played for the Jacksonville Jaguars.\\
- McCoy played for the Jacksonville Jaguars during his professional career.\\
\\
Please breakdown the following sentence into independent facts \\
        \bottomrule
    \end{tabular}
    \caption{
        A prompt given to InstructGPT to generate atomic facts for a given sentence. Model generated atomic facts were revised by human editors.
    }\label{tab:instruction-atomic-facts}
\end{table*}

\end{document}